\pdfoutput=1
\documentclass[11pt]{article}

\usepackage[final]{acl}

\usepackage{times}
\usepackage{latexsym}
\usepackage{amsmath}
\usepackage{amssymb}
\usepackage{enumitem}
\usepackage{booktabs}  % 导言区
\usepackage{multirow}  % 多行合并可选
\usepackage{array} 
\usepackage{makecell}
\usepackage{algorithm}
\usepackage{algpseudocode}

\usepackage[T1]{fontenc}
\usepackage[utf8]{inputenc}

\usepackage{microtype}

\usepackage{inconsolata}

\usepackage{graphicx}

\title{\textit{ReLoop}: "Seeing Twice and Thinking Backwards" via Closed-loop Training to Mitigate Hallucinations in Multimodal understanding}

\author{
  Jianjiang Yang\textsuperscript{1}\thanks{Equal contribution}, 
  Yanshu Li\textsuperscript{2}\footnotemark[1], 
  Ziyan Huang\textsuperscript{3}\footnotemark[1] \\
  \textsuperscript{1}University of Bristol, 
  \textsuperscript{2}Brown University, 
  \textsuperscript{3}South China University of Technology \\
  \texttt{edisonyang109@gmail.com}, \texttt{yanshu\_li1@brown.edu}, \texttt{bonnie.ziyan.huang@gmail.com}
}

\begin{document}
\maketitle
\begin{abstract}
While Multimodal Large Language Models (MLLMs) have achieved remarkable progress in open-ended visual question answering, they remain vulnerable to hallucinations. These are outputs that contradict or misrepresent input semantics, posing a critical challenge to the reliability and factual consistency. Existing methods often rely on external verification or post-hoc correction, lacking an internal mechanism to validate outputs directly during training. To bridge this gap, we propose \textbf{ReLoop}, a unified closed-loop training framework that encourages multimodal consistency for cross-modal understanding in MLLMs. ReLoop adopts a ring-shaped structure that integrates three complementary consistency feedback mechanisms, obliging MLLMs to \textbf{"seeing twice and thinking backwards"}. Specifically, ReLoop employs the frozen Consistency Feedback Plugin (CFP), comprising semantic reconstruction and visual description modules, along with an attention supervision module for attention alignment. These components collectively enforce semantic reversibility, visual consistency, and interpretable attention, enabling the model to correct its outputs during training. Extensive evaluations and analyses demonstrate the effectiveness of ReLoop in reducing hallucination rates across multiple benchmarks, establishing a robust method for hallucination mitigation in MLLMs. The code is available at: https://github.com/ZiyanHuang11/Reloop-hallucinations.

\end{abstract}

\section{Introduction}

\begin{figure}[ht]
\centering
\includegraphics[width=\linewidth]{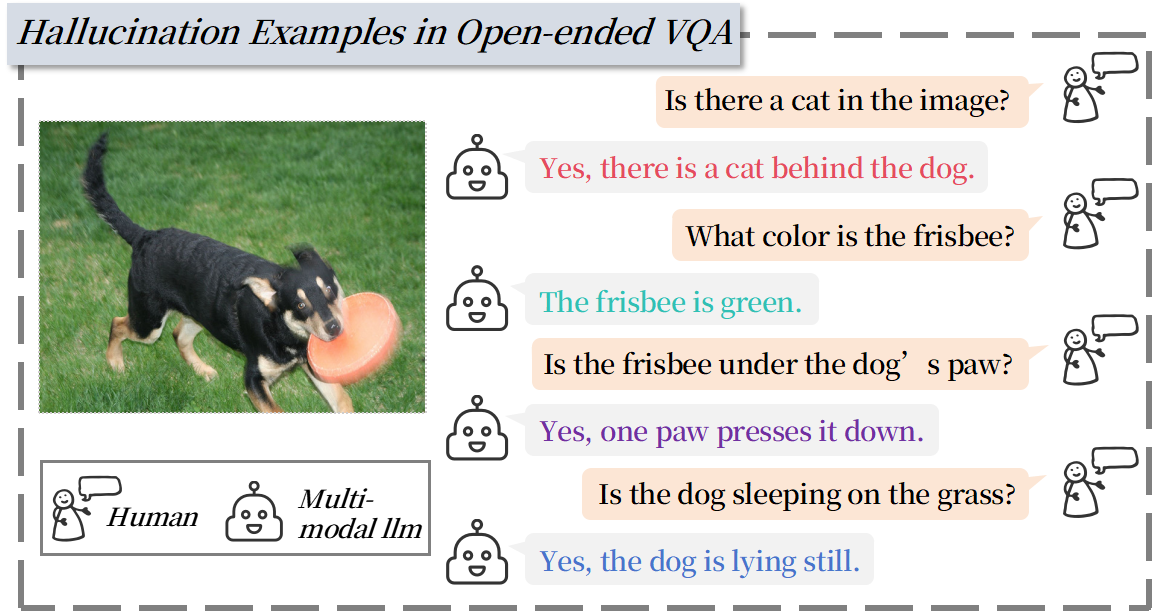}
\caption{Illustration of four major hallucination types in open-ended VQA.
Despite being visually grounded, MLLMs produce fluent but hallucinated responses across \textcolor{red!70!black}{object}, \textcolor{teal}{attribute}, \textcolor{violet}{relation}, and \textcolor{blue!70!black}{event} dimensions.} \label{fig:halluc_examples}
\end{figure}
In recent years, MLLMs~\citep{liu2023visual,openai2023gpt4v,li2023blip2} have demonstrated significant progress in bridging vision and language, addressing tasks such as visual question answering (VQA), image captioning, and instruction adherence. However, a fundamental difficulty that persists is hallucination, where the generation of outputs that are inconsistent with or unsupported by visual inputs~\citep{kalavasis2024limits}. Hallucinations are especially common in open-ended VQA circumstances, in which unclear or underspecified questions can result in factual mistakes. These hallucinations span diverse categories, including \textit{Object}, \textit{Attribute}, \textit{Relation}, and \textit{Event}. Figure~\ref{fig:halluc_examples} illustrates that a singular image of "a dog grasping an orange frisbee" can elicit various forms of hallucination: a fictitious "cat" (object), an incorrectly identified "green" frisbee (attribute), an erroneous spatial relation "under the paw" (relation), or a temporal misrepresentation "sleeping" (event). These errors are semantically plausible yet visually unfounded, posing major challenges for trustworthiness and safety of MLLMs across critical applications, including medical decision-making~\citep{kim2025medical}, robotic perception~\citep{park2023learning}, and autonomous navigation~\citep{alsulaimawi2025feedback}.

Existing works ~\citep{sun2023aligning,ayala2024reducing,sun2024redeep} often regard hallucination as an output-level anomaly that is corrected post hoc, overlooking its underlying cause. In practice, hallucinations frequently arise from misalignment between the input, visual content, and the model's latent reasoning. Without an internal supervision mechanism, models may produce fluent yet ungrounded answers. We argue that hallucination stems from the model's inability to validate its own output across modalities and recommend injecting this ability directly into training. 
% Rather than correcting errors post hoc, the training objective itself should guide the model to align its internal representations with the intended question and image semantics.

% To address this issue, 
We subsequently derive inspiration from human cognitive processes. When answering visual questions, individuals rarely rely on a single forward guess. Instead, after answering, they may reassess the question's intent, examine the visual scene, and refine conclusions—especially in the face of ambiguity or uncertainty. However, most models operate in a unidirectional manner, mapping $(Q, I \rightarrow A)$. As a result, once the model makes a prediction, there is no structured way to assess whether it actually understood the question, if the answer aligns with the visual evidence, or whether the model attended to the right regions in the image.

% To tackle this internal causes of hallucination
% To address this issue, we propose \textbf{ReLoop}, a cognitively inspired training framework that transforms MLLMs from static output generators into self-verifying systems. ReLoop introduces \textbf{a feedback-driven closed-loop supervision process}, enabling the model to revisit its predictions and validate their consistency with the original input. It guides the model to internalize semantic reversibility, aligning questions, answers, and visual context through structured reflection during training. In particular, ReLoop integrates two phases: a forward generation step that produces an answer from the image-question pair, and a feedback phase that revisits $(A, I)$ to recapture input semantics and assess internal consistency. This supervision is implemented via a \textbf{Consistency Feedback Plugin (CFP)}, comprising two frozen modules: (1) \textit{CFP-Lang} reconstructs the question $\hat{Q}^*$ from $(A, I)$ to supervise semantic alignment, and (2) \textit{CFP-Vis} generates a descriptive caption ${I}^*$ to assess factual grounding. In parallel, an \textbf{attention supervision module} extracts the model's token-to-image attention map $\mathcal{H}$ and compares it with an entropy-based pseudo-ground truth. All signals are integrated as differentiable losses in the overall optimization objective. This design encourages the model to "see twice and think backward"—first see to generate, see twice to reassess, and finally to adjust. 
To address this issue, we propose \textbf{ReLoop}, a cognitively inspired unified training framework that encourages multimodal consistency for cross-modal understanding in MLLMs. ReLoop implements a feedback-driven closed-loop supervision process, allowing the model to reassess its predictions and validate their consistency with the original input through multi-level supervision during training. Specifically, after MLLMs produce an answer from the image-question pair, Reloop enables the model to recapture input semantics and assess internal consistency via: a Consistency Feedback Plugin (CFP), comprising two frozen modules: (1) CFP-Lang reconstructs the question $\hat{Q}^*$ from $(A, I)$ to supervise semantic alignment, and (2) CFP-Vis generates a description ${I}^*$ to assess factual grounding. In parallel, an attention supervision module extracts the model's token-to-image attention map $\mathcal{H}$ and compares it with an entropy-based pseudo-ground truth. All signals are integrated as differentiable losses in the overall optimization objective. This design encourages the model to "see twice and think backward"—first look to answer ($Q,I \rightarrow A $), look twice to reassess ($A,I \rightarrow \hat{Q}^*, {I}^*, \mathcal{H}$), and finally to correct ($\hat{Q}^*, {I}^*, \mathcal{H}$ $\approx Q,I,\mathcal{H}_{\text{pseudo}}$).

ReLoop bridges the gap between perception and output. It turns the black-box understanding process into an interpretable, feedback-aware loop that continuously refines the model's internal representations. Our key contributions can be summarized clearly as follows:

\begin{itemize}[topsep=0em, itemsep=0em, leftmargin=1.2em]
\item We propose \textbf{ReLoop}, a cognitively inspired closed-loop training framework that ensures consistency among image, question, and answer modalities, effectively mitigating hallucinations in MLLMs.

\item We introduce three complementary consistency signals: semantic reconstruction, visual description, and attention alignment, to emulate the humanlike "reversible thinking" process and improve cross-modal consistency during training.

\item We provide a novel use of pretrained vision-language models by repositioning them as frozen Consistency Feedback Plugins (CFPs) in the training loop. Rather than functioning as typical forward-only encoders, they now perform in a reflective, backward supervisory role, producing feedback signals to guide the main model's alignment with multimodal semantics.
\end{itemize}
\begin{figure*}[t]
  \centering
  \includegraphics[width=\textwidth]{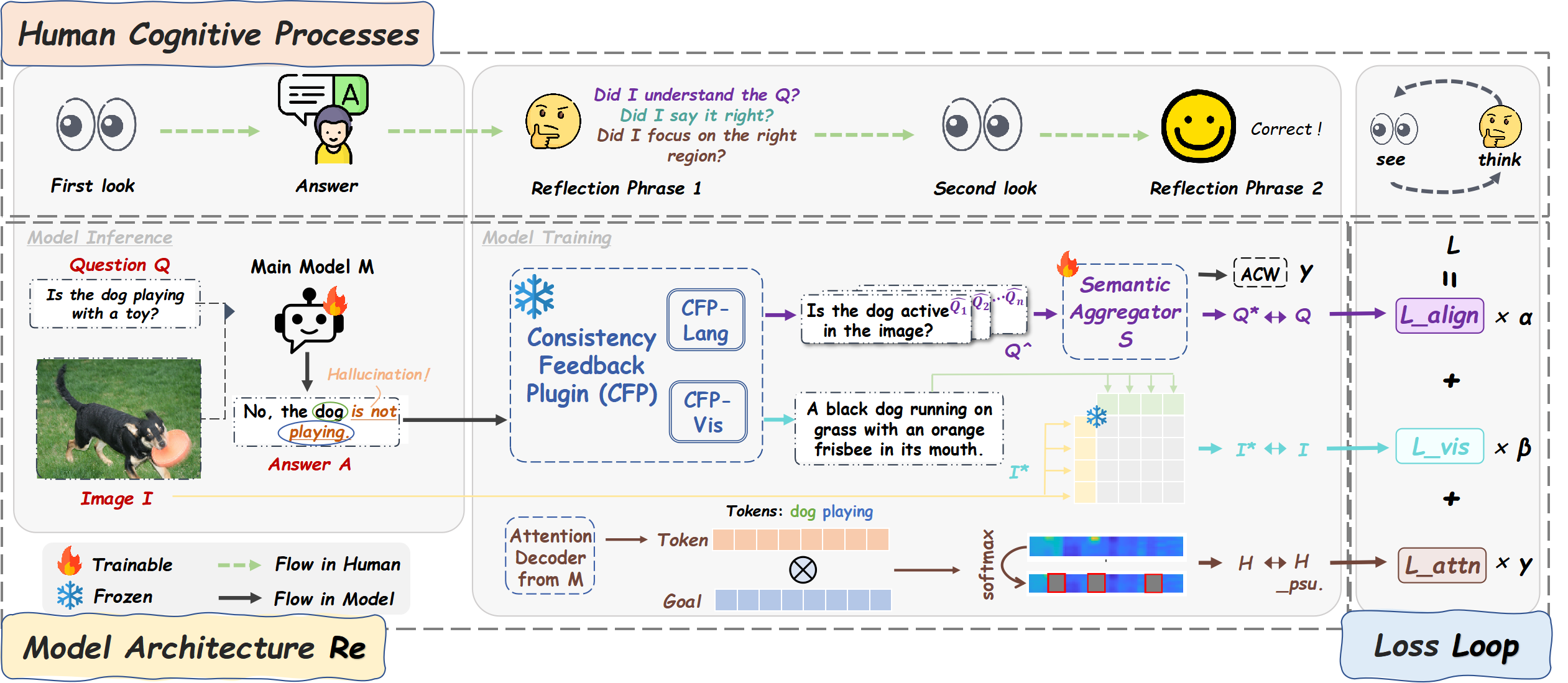}
  \caption{\textbf{Seeing Twice and Thinking Backwards: ReLooping Hallucination Suppression in Multimodal Language Models.} 
  This diagram aligns human cognitive phases (left) with model modules (right) in a closed-loop process. 
  The main model \(M\) produces an answer which is then introspected via CFP‑Lang (language reconstruction), CFP‑Vis (visual description), and internal cross‑attention maps. 
  Semantic aggregation, CLIP similarity, and entropy‑based soft masks produce feedback losses that are summed and back‑propagated to update \(M\) and the semantic aggregator \(S\).}
  \label{fig:reloop_arch}
\end{figure*}
%未完成
\section{Related Work}
\paragraph{Hallucination Mitigation in MLLMs.}
Multimodal LLMs frequently produce hallucinations: responses conflicting with visual inputs, such as inventing entities or misaligning semantics \citep{Li-hallucination-2023}. Recent mitigation efforts combine post-hoc correction and architectural refinement. Retrieval-augmented methods like \citep{mala2025hybrid} grounds outputs in external knowledge via hybrid retrievers, while \citep{ayala2024reducing} reduces hallucinations in structured outputs. Architectural solutions such as OPERA \citep{li2024opera} penalize over-trust during decoding, and preference-aligned training like TPO \citep{tpo2025} enhances vision grounding. Post-generation verification (Woodpecker \citep{woodpecker2023}) and decoding-time personalization (PAD \citep{chen2024pad}) complement training-time alignment; concurrent work improves multimodal ICL efficiency and task control \citep{li2025miv,li2025taco} and instruction-tuned dialog grounding \citep{luo2024nus}. Beyond images, video hallucination is diagnosed via fine-grained spatio–temporal grounding \citep{luo2025drvhierarchicalperceptiontemporalcognitionframework}.

% However, these approaches typically act as post-hoc add-ons and fail to embed semantic validation into the model's training dynamics.

\paragraph{Semantic Reversibility and Bidirectional Supervision.}
Human cognition leverages bidirectional reasoning to validate hypotheses: a principle termed "cognitive reversibility" \citep{johnson1983mental}. Recent works explore this idea through decoding-time strategies: Self-RAG \citep{asai2023selfrag} integrates retrieval-augmented generation with self-reflection, enabling models to critique and refine their outputs iteratively, while DeepSeekMath employs Group Relative Policy Optimization (GRPO) \citep{deepseek2024grpo}, enhancing mathematical reasoning by optimizing policy decisions based on group sampling strategies. Similarly, back-translation methods \citep{sennrich2016edinburgh} enforce answer-question consistency through round-trip translation. 

\paragraph{Cross-modal Consistency.}
Ensuring cross-modal consistency is vital for mitigating hallucinations in MLLMs. Recent methods enhance visual-text alignment to reduce semantic drift. VCD \citep{leng2023mitigating} contrasts outputs from original and perturbed images to promote grounding and reduce unimodal bias. HACL \citep{jiang2023hallucination} treats hallucinated captions as hard negatives to improve alignment. EAGLE \citep{villa2025eagle} further refines visual encoders post-pretraining, yielding better grounding and fewer hallucinations. Broader benchmark efforts catalog the MLLM evaluation landscape \citep{li2024survey}, including conversational aspect-based sentiment settings \citep{luo2024panosent}. 

% While prior work has addressed hallucination mitigation through external validation or decoding adjustments, ReLoop represents a novel direction by injecting semantic reversibility and multi-level supervision into the training process itself for better cross-modal consistency.
%未完成
\section{Preliminaries}

\subsection{Task Formulation: Open-ended Visual Question Answering}

We consider the task of open-ended VQA, where the model receives an image $I$ and a natural language question $Q$, and produces a free-form answer $A$. Unlike multiple-choice settings, this task requires the model to produce linguistically coherent and visually grounded responses without predefined options.

In this case, hallucination refers to answers that contradict the image $I$, misinterpret the question $Q$, or introduce unsupported content. 
% These errors undermine the model's reliability in applications that demand precise cross-modal understanding.

\subsection{Consistency Signals}

To encourage faithful understanding, we supervise the model using three types of cross-modal consistency signals:

\noindent\textbf{Linguistic Consistency.}
We verify whether the model's answer $A$ implies the same question intent as the original $Q$, by attempting to reconstruct $Q$ from $(A, I)$. This tests whether the model understood the question meaningfully.

\noindent\textbf{Visual Consistency.}
We evaluate whether the answer $A$ is factually grounded in image $I$, by generating a descriptive caption ${I}^*$ based on $(A, I)$ and checking its alignment with the image. This ensures that the response reflects the actual visual content.

\noindent\textbf{Attention Consistency.}
We examine whether the model attends to the correct regions of the image while producing $A$. This is assessed by comparing its internal attention map $\mathcal{H}$ with a soft pseudo-ground truth $\mathcal{H}_{\text{pseudo}}$ derived from entropy-based cues.

Together, these consistency signals serve as indirect evidence of whether the model truly grasps both the visual input and the question semantics.

\section{ReLoop Framework: Reflect, Recapture, and Optimize through a Closed-Loop Process}
We introduce \textbf{ReLoop}, a unified training framework aimed at reducing hallucinations in MLLMs for open-ended VQA answering. As illustrated in Figure~\ref{fig:reloop_arch}, the framework incorporates three complementary consistency feedback mechanisms: \textbf{semantic reconestruction}, \textbf{visual description}, and \textbf{attention alignment} to supervise the model toward producing answers faithful to both the question and the image.

These feedback signals are instantiated through a frozen \textbf{Consistency Feedback Plugin (CFP)}: semantic reconstruction (CFP-Lang) and visual description (CFP-Vis), and  \textbf{attention supervision} from the model itself. The CFP module is broadly compatible with a range of encoder-decoder or decoder-only MLLMs. During inference (\textit{First Look → Answer}), the model receives a question-image pair and produces an initial answer. The training process then begins with \textit{Reflect → Second Look → Correct}: the model examines its output through structured consistency feedback. Specifically, it "introspectively" asks:
\begin{itemize}[topsep=0em, itemsep=0em, leftmargin=1.2em]
  \item \textit{"Did I understand the $Q$?"} (→ semantic reconstruction)
  \item \textit{"Did I say it right?"} (→ visual description)
  \item \textit{"Did I focus on the right region?"} (→ attention alignment)
\end{itemize}

ReLoop decomposes hallucination mitigation into two interacting components:
\begin{itemize}[topsep=0em, itemsep=0em, leftmargin=1.2em]
    \item \textbf{\textit{"Re"}} emphasizes recapturing details, encouraging the model to reassess the semantic and visual cues from both question and image through CFP modules and token-level attention heatmaps.
    \item \textbf{\textit{"Loop"}} denotes a feedback-driven training loop. After each forward prediction, feedback from the three consistency pathways is aggregated into the loss function ($L_\text{align}$, $L_\text{vis}$, $L_\text{attn}$), driving iterative updates that refine the model's multimodal grounding and answer reliability.
\end{itemize}

% This "see → think → see again → correct" process constitutes a closed-loop optimization strategy that progressively enhances model understanding and reduces hallucinations.

\subsection{A Closed-loop Training }
The entire training process follows a closed-loop pattern, emulating "seeing twice and thinking backward". Each training step proceeds as follows:

\begin{enumerate}[topsep=0em, itemsep=0em, leftmargin=1.2em]
\item \textbf{First Look:} The main model $M$ takes the image $I$ and question $Q$ as input to produce an initial answer $A$.
\item \textbf{Reflect:} The model introspects on $A$ by reconstructing a proxy question $\hat{Q}$, generating a visual description ${I}^*$, and extracting token-level attention $\mathcal{H}$.
\item \textbf{Second Look:} The reconstructions are compared against the original inputs to compute consistency losses, capturing discrepancies in semantics, visual grounding, and attention focus.
\item \textbf{Correct:} All feedback signals are aggregated into $L_{\text{total}}$ to update $M$ and the semantic aggregator $S$ via backpropagation.
\end{enumerate}

This multi-stage loop is repeated across training epochs, leading to the model $M$ that gradually reduces hallucinations.

\subsection{Re: Recapturing Details for Consistency Supervision}
This stage corresponds to the training-time processes of "Reflect" and "Second Look", where the model reassesses its answers to recapture overlooked semantic and visual details. Three feedback pathways modules examine whether the model understood the question, correctly grounded its answer in the image, and attended to salient regions. 
\label{sec4.2}
\begin{figure*}[t]
\centering
\includegraphics[width=0.8\textwidth]{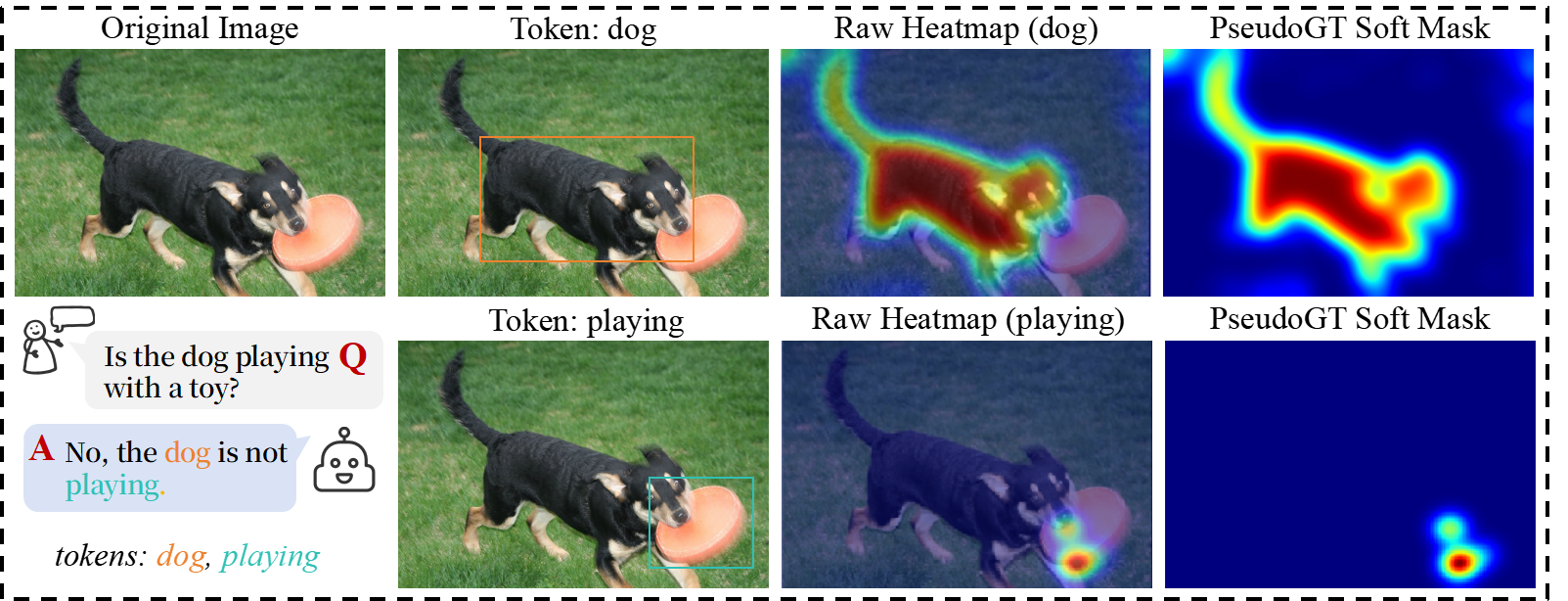}
\caption{
\textbf{Token-Level Attention Supervision.} Visualization of predicted attention $\mathcal{H}$ and entropy-based pseudo ground truth $\mathcal{H}_{\text{pseudo}}$ for two key answer tokens: \textit{dog} (top row) and \textit{playing} (bottom row). 
}
\label{fig:attn_supervision}
\end{figure*}
\subsubsection{CFP-lang: Language Reconstruction and Adaptive Consistency Weighting}
\label{sec:acw}
To evaluate whether the model correctly interprets the input question, we introduce a frozen language reconstruction module, CFP-lang. Given the answer-image pair $(A, I)$, CFP-lang produces a set of candidate reverse questions $\{ \hat{Q}_1, \hat{Q}_2, \ldots, \hat{Q}_k \}$ that approximate possible intents underlying the predicted answer. A lightweight semantic aggregator $S$, composed of a BERT encoder and a single-layer MLP, scores each candidate against the original question $Q$ using BERTScore. The highest-ranked proxy $\hat{Q}^*$ is selected to reflect the model's inferred intent.

However, directly enforcing alignment on all reconstructed questions may introduce noise, particularly when the produced answer is short or underspecified. To mitigate this, we introduce an Adaptive Consistency Weighting (ACW) mechanism, which adjusts the attention supervision (mentioned in section~\ref{sec:4.2.3}) strength based on the similarity between $Q$ and $\hat{Q}^*$:
\begin{equation}
\gamma = 
\begin{cases}
1.0 & \text{if BERTScore}(Q, \hat{Q}^*) \geq 0.8 \\
0.1 & \text{if } 0.6 \leq \text{BERTScore}(Q, \hat{Q}^*) < 0.8 \\
0.01 & \text{if BERTScore}(Q, \hat{Q}^*) < 0.6
\end{cases}
\label{equ:gama}
\end{equation}
Rather than discarding low-confidence pairs, this soft weighting ensures that stronger semantic matches contribute more prominently to the learning objective. The language consistency loss is computed as:
\begin{equation}
L_{\text{align}} = 1 - \text{BERTScore}(Q, \hat{Q}^*)
\end{equation}
\subsubsection{CFP-visual: Visual Description and Similarity Supervision}
To validate whether the produced answer $A$ is visually grounded in the image $I$, we employ a frozen visual description module, CFP-visual.  Given $(A, I)$, it generates a caption ${I}^*$ describing the image content implied by the answer. We then compute the cosine similarity between the CLIP-encoded vectors of $I$ and ${I}^*$, and derive the visual consistency loss as:
\begin{equation}
L_{\text{vis}} = 1 - \cos(\text{CLIP}_{\text{img}}(I), \text{CLIP}_{\text{text}}({I}^*))
\end{equation}
% This encourages the model to produce answers that are not only linguistically valid, but also factually coherent with the image content.
\subsubsection{Attention Supervision via Heatmap Consistency}
\label{sec:4.2.3}
To enhance interpretability and mitigate hallucinations arising from inattentive or unstable decoding, we explicitly supervise the model's token-level cross-attention patterns.  From the decoder of the main model $M$, we extract attention maps $\mathcal{H}$, which indicate the spatial focus during answer generation. We construct a soft pseudo-ground-truth heatmap $\mathcal{H}_{\text{pseudo}}$ using entropy-based masking (Detailed explanation can be found in Appendix~\ref{appendixC}). This method preserves uncertainty information and avoids brittle hard labels. As illustrated in Figure~\ref{fig:attn_supervision}, well-grounded tokens (e.g., \textit{dog}) yield concentrated heatmaps aligned with visual evidence, while hallucinated tokens (e.g., \textit{playing}) produce offset patterns.
\begin{table*}[t]
\centering
\renewcommand{\arraystretch}{1.2}
\small
\setlength{\tabcolsep}{0pt}
\begin{tabular}{
  >{\centering\arraybackslash}p{1.8cm}  % Type
  >{\centering\arraybackslash}p{1.8cm}  % Module
  >{\centering\arraybackslash}p{2.2cm}  % Signal Type
  >{\centering\arraybackslash}p{2cm}    % Baseline
  >{\centering\arraybackslash}p{2cm}    % ReLoop
  >{\centering\arraybackslash}p{1.6cm}  % Delta Mean
  >{\centering\arraybackslash}p{1.4cm}  % Baseline Hallu
  >{\centering\arraybackslash}p{1.4cm}  % ReLoop Hallu
  >{\centering\arraybackslash}p{1.4cm}  % Delta Rate
}
\toprule
\textbf{Type} & \textbf{Module} & \textbf{Signal Type} & \textbf{Baseline} & \textbf{ReLoop} & \textbf{$\Delta$Mean} & \textbf{Baseline Hallu.} & \textbf{ReLoop Hallu.} & \textbf{$\Delta$Rate} \\
\midrule
\multirow{3}{*}{Object}
  & Visual    & CLIP($I$, ${I}^*$)     & 28.02 $\pm$ 3.10 & 29.46 $\pm$ 3.27 & $\uparrow$1.44 & \multirow{3}{*}{24.5\%} & \multirow{3}{*}{10.3\%} & \multirow{3}{*}{$\downarrow$14.2\%} \\
  & Language  & BERT($Q$, $\hat{Q}$)     & 0.862 $\pm$ 0.022 & 0.873 $\pm$ 0.024 & $\uparrow$0.011 & & & \\
  & Attention & Entropy($\mathcal{H}$)  & 1.31 $\pm$ 0.40   & 1.28 $\pm$ 0.45   & $\downarrow$0.03 & & & \\
\midrule
\multirow{3}{*}{Attribute}
  & Visual    & CLIP($I$, ${I}^*$)     & 26.59 $\pm$ 3.31 & 26.81 $\pm$ 3.41 & $\uparrow$0.22 & \multirow{3}{*}{7.3\%} & \multirow{3}{*}{4.0\%} & \multirow{3}{*}{$\downarrow$3.3\%} \\
  & Language  & BERT($Q$, $\hat{Q}$)     & 0.868 $\pm$ 0.025 & 0.894 $\pm$ 0.028 & $\uparrow$0.026 & & & \\
  & Attention & Entropy($\mathcal{H}$)  & 1.36 $\pm$ 0.46   & 1.32 $\pm$ 0.52   & $\downarrow$0.04 & & & \\
\midrule
\multirow{3}{*}{Relation}
  & Visual    & CLIP($I$, ${I}^*$)     & 27.22 $\pm$ 3.26 & 28.01 $\pm$ 3.38 & $\uparrow$0.79 & \multirow{3}{*}{13.2\%} & \multirow{3}{*}{7.6\%} & \multirow{3}{*}{$\downarrow$5.6\%} \\
  & Language  & BERT($Q$, $\hat{Q}$)     & 0.855 $\pm$ 0.020 & 0.875 $\pm$ 0.023 & $\uparrow$0.020 & & & \\
  & Attention & Entropy($\mathcal{H}$)  & 1.39 $\pm$ 0.43   & 1.34 $\pm$ 0.50   & $\downarrow$0.05 & & & \\
\midrule
\multirow{3}{*}{Event}
  & Visual    & CLIP($I$, ${I}^*$)     & 26.63 $\pm$ 3.08 & 26.94 $\pm$ 3.37 & $\uparrow$0.31 & \multirow{3}{*}{10.4\%} & \multirow{3}{*}{5.2\%} & \multirow{3}{*}{$\downarrow$5.2\%} \\
  & Language  & BERT($Q$, $\hat{Q}$)     & 0.861 $\pm$ 0.024 & 0.877 $\pm$ 0.029 & $\uparrow$0.016 & & & \\
  & Attention & Entropy($\mathcal{H}$)  & 1.33 $\pm$ 0.42   & 1.51 $\pm$ 0.55   & $\uparrow$0.18 & & & \\
\bottomrule
\end{tabular}
\caption{Effect of ReLoop on consistency and hallucination reduction across different hallucination types. We compare MiniGPT-4 (baseline) and ReLoop in terms of signal outputs from three frozen feedback modules: visual grounding (CLIP similarity), semantic alignment (BERTScore), and attention focus (entropy). $\Delta$ denotes the absolute change in signal quality after applying ReLoop.}
\label{tab:hallu-type-full}
\end{table*}
We enforce alignment between $\mathcal{H}$ and $\mathcal{H}_{\text{pseudo}}$ by minimizing the KL divergence:
\begin{equation}
L_{\text{attn}} = \text{KL}(\mathcal{H} \parallel \mathcal{H}_{\text{pseudo}})
\end{equation}
 
% This encourages the model to attend to semantically relevant regions, reducing hallucinations caused by unstable or inattentive decoding behavior.
\subsection{Loop: Feedback Aggregation, Alignment, and Optimization}
After consistency signals are computed from language, vision, and attention supervision, ReLoop aggregates them into a unified training objective. This stage corresponds to the "Correction" step in the loop, where the model updates its parameters based on multi-perspective feedback. The total loss combines standard supervision with the three consistency terms:
\label{sec4.3}
\begin{equation}
L_{\text{total}} = L_{\text{sft}} + \alpha \cdot L_{\text{align}} + \beta \cdot L_{\text{vis}} + \gamma \cdot L_{\text{attn}} + \lambda \cdot \Omega(\theta)
\end{equation}
where $L_{\text{sft}}$ is the token-level cross-entropy loss, and $\Omega(\theta)$ is an L2 regularization term. The consistency weights are empirically set as $\alpha=1.0$, $\beta=0.7$, $\lambda=10^{-5}$ and $\gamma$ is defined in Equation~\ref{equ:gama}.

Only the parameters of the main model $M$ and the semantic aggregator $S$ are updated during training. All feedback modules, including CFP-Lang, CFP-Vis, attention supervision, and CLIP, remain frozen. 

\section{Experimental Setup}
\noindent\textbf{Training Data.}
We curate 30K high-quality $\{$$I,Q,A$$\}$ from LLaVA-Instruct-150K. To simulate hallucination supervision, we generate contrastive examples by perturbing key semantics (e.g., \textit{objects, attributes, relations, event}), enabling fine-grained control over hallucination types. Details can be found in Appendix~\ref{a1}~\ref{sec:appendix_dataset}.

\begin{figure}[t]
  \centering
  \includegraphics[width=0.9\columnwidth]{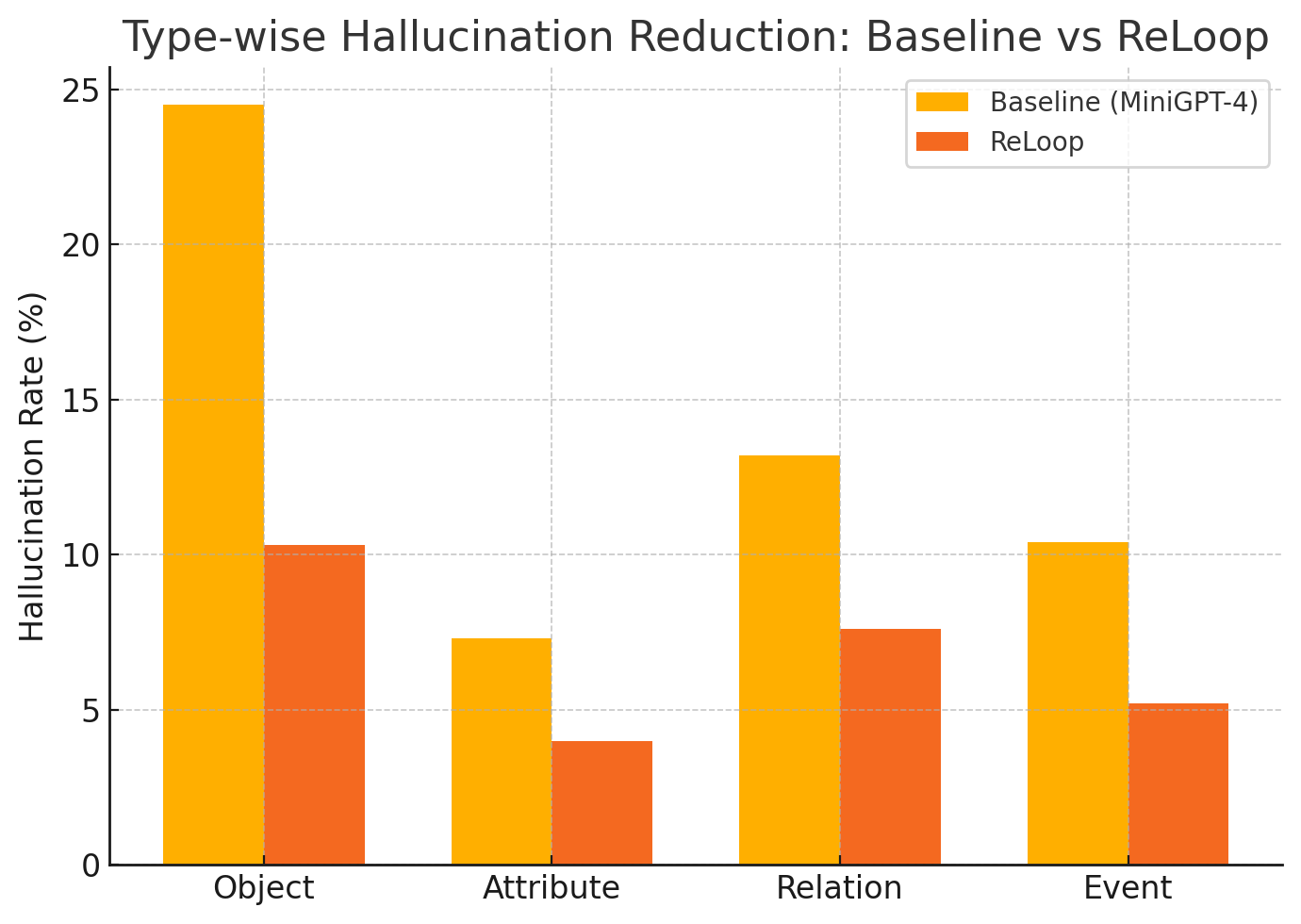}
  \caption{Type‑wise hallucination rates (\%) for baseline (MiniGPT-4) and ReLoop models.}
  \label{fig:halluc_reduction}
\end{figure}

\noindent\textbf{Evaluation Benchmarks and Metrics.}
We evaluate ReLoop on a broad range of hallucination and multimodal understanding benchmarks, including POPE~\citep{Li-hallucination-2023}, CHAIR~\citep{rohrbach2018object}, AMBER~\citep{wang2023llm}, MMHal-B~\citep{sun2023aligning}, HallusionBench~\citep{Guan_2024_CVPR}, Faith/FaithS~\citep{jing-etal-2024-faithscore}, and MME~\citep{fu2023mme}. Full definitions can be found in Appendix~\ref{sec:appendix_metrics}~\ref{a4}.

\begin{figure*}[t]
\centering
\includegraphics[width=0.9\textwidth]{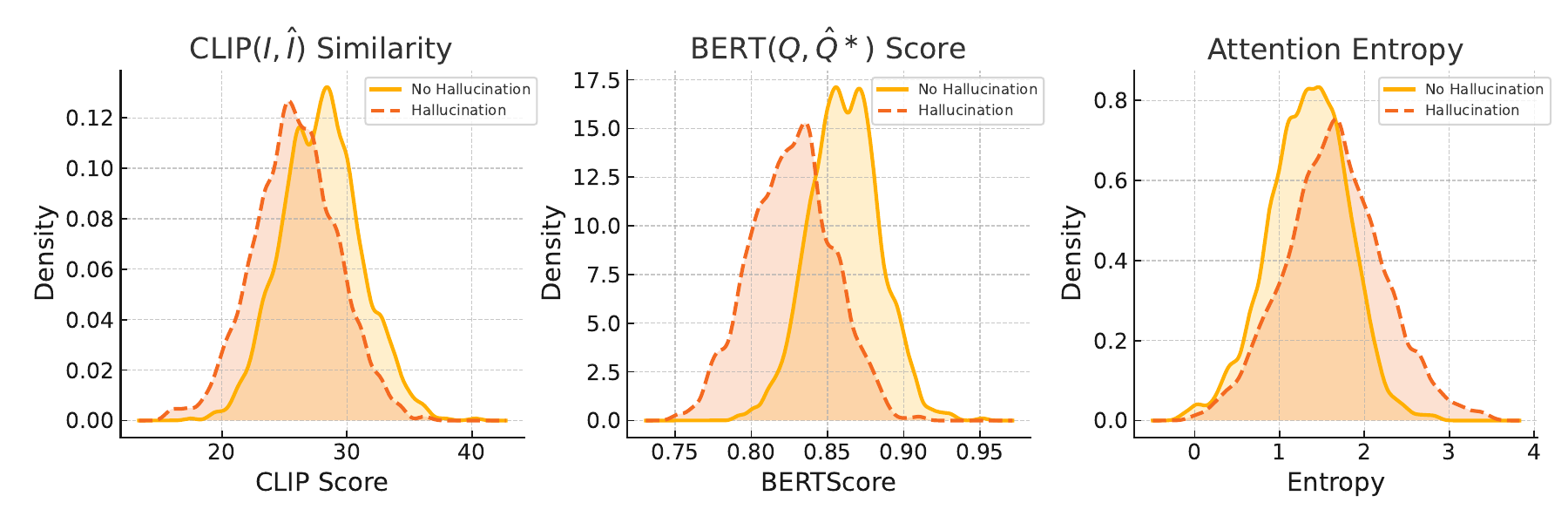} 
\caption{KDE distributions of CLIP similarity, BERTScore, and attention entropy for hallucinated and non-hallucinated samples. ReLoop's frozen modules exhibit sharp signal shifts that serve as reliable supervision sources.}
\label{fig:kde}
\end{figure*}

\noindent\textbf{Baselines.}
We use MiniGPT-4 as the baseline model in Experiment~\ref{6.1} and compare against LLaVA-1.5 variants trained with LLaVA-RLHF~\cite{sun2023aligning}, HA-DPO~\cite{zhao2023beyond}, POVID~\cite{zhou2024povid}, and Visual Contrastive Decoding (VCD)~\cite{damonlpsg2023vcd}. 
For robustness analyses, we adopt \emph{LLaVA-1.5 + ReLoop} as the canonical setting and report stress tests under noisy external supervision and nonsensical answers (Table~\ref{tab:robustness_combined}). 
Unless otherwise specified, all baselines share the same backbone, data, and training protocol for a fair comparison. Implementation details are provided in Appendix~\ref{sec:appendix_alignment}.

% \begin{table}[t]
% \centering
% \setlength{\tabcolsep}{1pt}
% \small
% \begin{tabular}{@{}lccccc@{}}
% \toprule
% \textbf{Module} & \textbf{Signal Type} & \textbf{Hal.} & \textbf{Mean ± Std} & $\Delta$Mean \\
% \midrule
% \multirow{2}{*}{Visual}  
%     & \multirow{2}{*}{\makecell[c]{CLIP($I$, ${I}^*$)\\cosine}} & No  & 27.89 ± 3.21 & — \\
%     &                             & Yes & 25.64 ± 3.34 & ↓ 2.25 \\
% \midrule
% \multirow{2}{*}{Langu.} 
%     & \multirow{2}{*}{\makecell[c]{BERT($Q$, $\hat{Q}$)\\cosine}} & No  & 0.861 ± 0.023 & — \\
%     &                             & Yes & 0.827 ± 0.026 & ↓ 0.034 \\
% \midrule
% \multirow{2}{*}{Atten.} 
%     & \multirow{2}{*}{\makecell[c]{Entropy\\($\mathcal{H}$)}}      & No  & 1.36 ± 0.47   & — \\
%     &                             & Yes & 1.67 ± 0.58   & ↑ 0.31 \\
% \bottomrule
% \end{tabular}

% \caption{Mean ± standard deviation of signal outputs under hallucinated and non-hallucinated conditions. Hal. indicates whether the objects are hallucinated. Hallucinated cases exhibit lower CLIP and BERT alignment, and higher attention entropy. All differences are statistically significant ($p < 10^{-5}$).}
% \label{tab:signal_stats}
% \end{table}

\begin{table*}[t]
\centering
\renewcommand{\arraystretch}{0.9}
\small
\setlength{\tabcolsep}{2pt} % 控制列间距
\begin{tabular}{
  >{\centering\arraybackslash}p{3.3cm}  % Model
  >{\centering\arraybackslash}p{1.7cm}  % POPE
  >{\centering\arraybackslash}p{1.7cm}  % CHAIR_s
  >{\centering\arraybackslash}p{1.7cm}  % CHAIR_i
  >{\centering\arraybackslash}p{1.7cm}  % F1
  >{\centering\arraybackslash}p{1.7cm}  % Faith
  >{\centering\arraybackslash}p{1.7cm}  % FaithS
}
\toprule
\multirow{2}{*}{\textbf{Model}} 
& \multicolumn{3}{c}{\textbf{Hallucination Suppression}} 
& \multicolumn{3}{c}{\textbf{Cross-modal Faithfulness}} \\
\cmidrule(lr){2-4} \cmidrule(lr){5-7}
& POPE$\uparrow$ & CHAIR$_s\downarrow$ & CHAIR$_i\downarrow$ 
& F1$\uparrow$ & Faith$\uparrow$ & FaithS$\uparrow$ \\
\midrule
MiniGPT-4              & 82.3 & 49.0 & 22.7 & 63.2 & 86.7 & 68.5 \\
\quad + ReLoop         & \textbf{83.9} & \textbf{38.8} & \textbf{20.5} & \textbf{69.9} & \textbf{88.6} & \textbf{71.3} \\
\midrule
InstructBLIP           & 83.8 & 47.8 & 20.6 & \textbf{68.4} & 87.3 & 69.8 \\
\quad + ReLoop         & \textbf{85.3} & \textbf{36.9} & \textbf{17.5} & 67.0 & \textbf{88.5} & \textbf{73.2} \\
\midrule
LLaVA-1.5     & 85.7 & 53.5 & 24.2 & 65.8 & \textbf{89.5} & \textbf{75.8} \\
\quad + ReLoop         & \textbf{86.3} & \textbf{40.2} & \textbf{16.2} & \textbf{70.3} & 89.2 & 75.3 \\
\midrule
LLaVA-1.6     & 86.8 & 52.0 & 21.8 & 67.3 & \textbf{89.4} & \textbf{76.6} \\
\quad + ReLoop         & \textbf{87.9} & \textbf{38.5} & \textbf{16.1} & \textbf{71.1} & 89.2 & 76.2 \\
\midrule
Qwen-VL-2.5   & 89.3 & 47.3 & 20.8 & 69.1 & 89.7 & 76.5 \\
\quad + ReLoop         & \textbf{90.7} & \textbf{37.6} & \textbf{16.6} & \textbf{72.5} & \textbf{90.4} & \textbf{77.8} \\
\midrule
mPLUG-owl              & 89.1 & 62.5 & 31.0 & 58.9 & \textbf{88.3} & \textbf{72.7} \\
\quad + ReLoop         & \textbf{90.9} & \textbf{42.5} & \textbf{21.8} & \textbf{66.5} & 87.9 & 71.0 \\
\midrule
ShareGPT4V             & 88.2 & 50.2 & 21.8 & 68.0 & 88.2 & 73.6 \\
\quad + ReLoop         & \textbf{89.7} & \textbf{44.9} & \textbf{21.5} & \textbf{69.2} & \textbf{89.3} & \textbf{74.8} \\
\bottomrule
\end{tabular}
\caption{Performance comparison of various LVLMs with and without ReLoop. Hallucination is measured by POPE, CHAIR$_s$, and CHAIR$_i$; cross-modal faithfulness is evaluated using F1, Faith, and FaithS. ↓ indicates lower is better; ↑ indicates higher is better.}
\label{tab:model_performance}
\end{table*}

\section{Results and Analysis}
% In this section, our analysis specifically targets (1) why hallucinations occur (internal representation gaps), (2) how effectively structured feedback mechanisms in Reloop address these gaps, and (3) to what extent these improvements generalize across diverse evaluation scenarios.
% we systematically investigate the internal factors contributing to hallucinations in MLLMs, and evaluate how structured multimodal consistency feedback mechanisms introduced by ReLoop effectively mitigate these issues. 
\subsection{Identify Internal Causes of Hallucinations: Module Signals vs. Hallucination States}
\label{6.1}
We first aim to pinpoint internal representation deficiencies that drive hallucination behaviors across different hallucination types. We analyze consistency signal deviations produced by ReLoop's frozen supervision modules, with hallucinated versus non-hallucinated samples. Responding: \textit{"Did I understand the question?”} (language, via BERTScore); \textit{"Did I say it right?”} (visual, via CLIP similarity); \textit{"Did I focus on the right region?”} (attention, via entropy).

\noindent\textbf{Multimodal hallucinations stem from structured, modality-specific representation gaps.}  
As shown in Table~\ref{tab:hallu-type-full}, hallucinated responses are consistently associated with lower CLIP similarity (–2.25), reduced BERTScore (–0.034), and higher attention entropy (+0.31). Figure~\ref{fig:kde} reveals distinct signal patterns associated with different hallucination types. Object hallucinations correspond to a clear leftward shift in CLIP similarity, indicating weaker visual grounding. Attribute hallucinations are marked by lower BERTScore, reflecting reduced semantic alignment. Event hallucinations show higher attention entropy, suggesting that the model distributes focus more broadly, which may help in capturing complex scenes but also increases the risk of focusing on irrelevant regions.

\noindent\textbf{Signal dynamics vary by hallucination type.}  
(1) \textit{Object hallucinations} are primarily rooted in the visual module. They often manifest as hallucinated entities not present in the image. ReLoop yields a significant gain in CLIP similarity (↑1.44) and a decrease in attention entropy (↓0.03), suggesting enhanced image-text alignment and focused visual grounding.  
(2) \textit{Attribute hallucinations} show the largest improvement in BERTScore (↑0.026) and only a slight change in CLIP similarity (↑0.22), indicating that semantic reconstruction plays a more important role than visual grounding. This aligns with their nature: attributes often relate to textual misinterpretation (e.g., \textit{color or size}), even when visual cues are present.  
(3) \textit{Relation hallucinations} involve complex spatial or relational semantics and display moderate improvements across all three signals (CLIP↑0.79, BERT↑0.020, Entropy↓0.05), suggesting that ReLoop's multi-signal supervision addresses cross-modal misalignment collaboratively.  
(4) \textit{Event hallucinations} are primarily tied to attention misallocation. ReLoop improves CLIP (↑0.31) and BERT (↑0.016) slightly, but entropy increases (↑0.18), reflecting broader attention scopes. This likely helps avoid fixation on irrelevant regions, especially in dynamic or temporally inferred scenes. Figure~\ref{fig:halluc_reduction} shows that ReLoop successfully mitigates hallucinations compared to MiniGPT-4 across four hallucination types.
\begin{table*}[t]
\centering
\renewcommand{\arraystretch}{0.9}
\small
\setlength{\tabcolsep}{2pt}
\begin{tabular}{
  >{\centering\arraybackslash}p{3.8cm}  % Ablation Version
  >{\centering\arraybackslash}p{1.7cm}  % POPE
  >{\centering\arraybackslash}p{1.7cm}  % CHAIR_s
  >{\centering\arraybackslash}p{1.7cm}  % CHAIR_i
  >{\centering\arraybackslash}p{1.7cm}  % F1
  >{\centering\arraybackslash}p{1.7cm}  % Faith
  >{\centering\arraybackslash}p{1.7cm}  % FaithS
}
\toprule
\multirow{2}{*}{\textbf{Ablation Version}} 
& \multicolumn{3}{c}{\textbf{Hallucination Suppression}} 
& \multicolumn{3}{c}{\textbf{Cross-modal Faithfulness}} \\
\cmidrule(lr){2-4} \cmidrule(lr){5-7}
& POPE$\uparrow$ & CHAIR$_s\downarrow$ & CHAIR$_i\downarrow$ 
& F1$\uparrow$ & Faith$\uparrow$ & FaithS$\uparrow$ \\
\midrule
MiniGPT-4 & 83.0 & 49.0 & 22.7 & 60.2 & 84.3 & 64.2 \\
w/o Consistency Supervision & 84.2 & 47.4 & 21.6 & 60.7 & 86.7 & 68.5 \\
w/o Gating \& Aggregator & \textbf{85.4} & 39.8 & 19.7 & 60.4 & 88.1 & 71.6 \\
w/o Attention Supervision & 83.6 & 40.2 & 20.1 & 61.9 & 86.3 & 67.5 \\
Full ReLoop & 84.9 & \textbf{38.3} & \textbf{18.9} & \textbf{63.1} & \textbf{88.6} & \textbf{72.8} \\
\bottomrule
\end{tabular}
\caption{Performance comparison of ReLoop under different ablation configurations on MiniGPT-4. Removing consistency supervision results in the worst faithfulness and hallucination rate, while full ReLoop delivers the best overall performance. Although gating removal slightly improves POPE, it hurts precision (F1) and consistency.}
\label{tab:ablation}
\end{table*}
\begin{table*}[htbp]
\centering
\renewcommand{\arraystretch}{0.9}
\small
\setlength{\tabcolsep}{2pt}
\begin{tabular}{
  >{\centering\arraybackslash}p{3.8cm}  % Method
  >{\centering\arraybackslash}p{1.7cm}  % POPE
  >{\centering\arraybackslash}p{1.7cm}  % CHAIR_s
  >{\centering\arraybackslash}p{1.7cm}  % CHAIR_i
  >{\centering\arraybackslash}p{1.7cm}  % F1
  >{\centering\arraybackslash}p{1.7cm}  % Faith
  >{\centering\arraybackslash}p{1.7cm}  % FaithS
}
\toprule
\multirow{2}{*}{\textbf{Method}} 
& \multicolumn{3}{c}{\textbf{Hallucination Suppression}} 
& \multicolumn{3}{c}{\textbf{Cross-modal Faithfulness}} \\
\cmidrule(lr){2-4} \cmidrule(lr){5-7}
& POPE$\uparrow$ & CHAIR$_s\downarrow$ & CHAIR$_i\downarrow$ 
& F1$\uparrow$ & Faith$\uparrow$ & FaithS$\uparrow$ \\
\midrule
LLaVA-1.5              & 83.5 & 53.9 & 23.5 & 63.2 & 86.9 & 70.5 \\
+ LLaVA-RLHF           & \textbf{88.2} & 44.5 & \underline{20.1} & \underline{67.0} & \underline{89.0} & \underline{74.4} \\
+ HA-DPO               & 86.7 & 52.3 & 21.6 & 65.4 & 88.4 & 73.5 \\
+ POVID                & 84.3 & 53.2 & 24.2 & 64.7 & 87.3 & 71.8 \\
+ VCD         & 86.8 & \underline{43.1} & 20.2 & 66.9 & 88.8 & 73.6 \\
+ \textbf{ReLoop}      & \underline{87.9} & \textbf{42.0} & \textbf{19.5} & \textbf{67.4} & \textbf{89.5} & \textbf{75.1} \\
\bottomrule
\end{tabular}
\caption{Performance comparison of ReLoop with alignment-enhancing baselines for LLaVA-1.5 on hallucination suppression and cross-modal faithfulness. Best scores are in \textbf{bold} and the second best are \underline{underlined}.}
\label{tab:alignment_finegrained}
\end{table*}

\subsection{Effects of Structured Feedback in ReLoop}
\label{6.2}
Motivated by earlier findings, we evaluate how effectively ReLoop's structured feedback enhances semantic grounding across five representative LVLMs (Table~\ref{tab:model_performance}). The observed improvements span models with diverse architectures and training paradigms, showing that ReLoop is broadly compatible and easily integrable into various LVLMs.

\noindent\textbf{Hallucination Suppression.} ReLoop significantly reduces references to non-existent entities. InstructBLIP shows 22.8\%/15.0\% reductions. LLaVA-1.5 improves by 24.9\%/33.1\%, and strong backbones exhibit the same trend: LLaVA-1.6 achieves $\sim$26\%/$\sim$26\% drops, while Qwen-VL-2.5 yields $\sim$20.5\%/$\sim$20.2\%. Similar effects hold for mPLUG-owl and ShareGPT4V. These reductions confirm that ReLoop enhances visual grounding and spatial precision across backbones.

\noindent\textbf{Cross-modal Faithfulness.} ReLoop also enhances cross-modal faithfulness. F1 increases on MiniGPT-4 (+10.6\%), LLaVA-1.5 (+6.8\%), LLaVA-1.6 ($\sim$5.6\%), and Qwen-VL-2.5($\sim$4.9\%); InstructBLIP maintains comparable F1 while gaining on faith metrics. FaithS improves for MiniGPT-4 (+2.8), InstructBLIP (+3.4), Qwen-VL-2.5 (+1.3), and ShareGPT4V (+1.2), and remains near-parity on LLaVA-1.6. These gains suggest that the model not only grounds responses more accurately in the image but also maintains semantic alignment with the question intent. 

\subsection{Robustness Under Noisy Supervision}
\label{sec:robustness-noise}
We next ask whether closed-loop training remains stable when external supervision is imperfect or when the initial answer signal is degenerate. We stress-test ReLoop under (i) noisy teacher feedback on the visual channel and (ii) nonsensical answers that could mislead the loop.

\noindent\textbf{Closed-loop supervision is resilient to noisy teachers.}
We corrupt 15\% of visual descriptions fed to the CFP-Vis branch with pseudo-random text. As shown in Table~\ref{tab:robustness_combined}, the average CLIP similarity decreases ($\Delta$CLIP $=-0.11$), yet core hallucination metrics remain stable (POPE $87.9\!\rightarrow\!86.8$, CHAIR-s $42.0\!\rightarrow\!42.6$, CHAIR-i $19.5\!\rightarrow\!20.2$). This suggests that multi-signal aggregation dilutes spurious teacher cues, preserving cross-modal consistency even when the visual supervisor is noisy.
\begin{table*}[t]
\centering
\small
\setlength{\tabcolsep}{2pt}
\begin{tabular}{ccccccc}
\toprule
\textbf{Setting} & \textbf{POPE}$\uparrow$ & \textbf{CHAIR-s}$\downarrow$ & \textbf{CHAIR-i}$\downarrow$ & \textbf{$\Delta$CLIP-sim} ($\uparrow$) & \textbf{$\gamma$ dist. (1.0/0.1/0.01)} & \textbf{$\Delta$BERTScore} ($\uparrow$) \\
\midrule
LLaVA-1.5 + ReLoop (clean) & 87.9 & 42.0 & 19.5 & ---    & 52/41/9  & --- \\
+ Teacher noise (15\%)     & 86.8 & 42.6 & 20.2 & $-0.11$ & ---       & --- \\
+ Answer noise (15\%)      & 87.2 & 43.1 & 20.4 & ---     & 35/43/22  & $-0.06$ \\
\bottomrule
\end{tabular}
\caption{Robustness under noisy supervision. Teacher-side visual description corruption (15\%) and answer-side nonsense injection (15\%) have limited impact on core hallucination metrics. CLIP similarity drops under teacher noise, whereas ACW re-allocates per-sample weights under answer noise (smaller high-confidence mass).}
\label{tab:robustness_combined}
\end{table*}

\noindent\textbf{ACW suppresses nonsensical answers without destabilizing training.}
We replace 15\% of answers with meaningless strings while keeping ACW active. Table~\ref{tab:robustness_combined} shows a predictable reallocation of per-sample weights: the mass at high confidence shrinks ($\gamma{=}1.0$: $52\%\!\rightarrow\!35\%$), medium/low weights grow ($0.1/0.01$: $41/9\%\!\rightarrow\!43/22\%$), and semantic alignment only mildly drops ($\Delta$BERTScore$=-0.06$), while POPE/CHAIR remain essentially unchanged. This confirms that ACW down-weights misleading answer signals before they influence learning.

\noindent\textbf{Natural noise robustness via fourfold filtering.}
ReLoop’s robustness emerges from a fourfold filter: three orthogonal supervision signals (language via BERTScore, visual via CLIP, attention via entropy-aware $\mathcal{H}_{\text{pseudo}}$) plus ACW’s discrete gating $\gamma\!\in\!\{1,0.1,0.01\}$ (Sec.~\ref{sec4.2}). Noisy teacher feedback is first cross-validated across modalities, then attenuated by ACW, and finally diluted in the multi-loss objective (Sec.~\ref{sec4.3}), so that biased cues have limited influence on the update direction. Empirically, the stability of POPE/CHAIR under teacher-side corruption and the expected shift in the $\gamma$ distribution under answer nonsense together indicate that the closed loop self-regularizes and converges stably despite imperfect teachers.

\subsection{Ablation Study}
To assess the contribution of each component in ReLoop, we perform a coarse-grained ablation study over four configurations (Table~\ref{tab:ablation}). Removing consistency supervision leads to the highest hallucination rates (CHAIR$_s$: 47.4) and lowest semantic faithfulness (FaithS: 68.5), highlighting its central role. Attention supervision also proves important, as its removal moderately reduces FaithS. While removing gating slightly improves POPE, it harms F1 and hallucination suppression. Full ReLoop achieves the best overall results, reducing CHAIR$_s$ by 10.7 and increasing FaithS by 8.6 over the baseline. These findings underscore the complementary roles of all modules and the importance of structured feedback for robust alignment.

% \begin{figure}[t]
%   \centering
%   \includegraphics[width=\columnwidth]{fig/ablation_dual_trend.png}
%   \caption{Dual-axis analysis of GPTScore and hallucination rate across ReLoop ablation settings.}
%   \label{fig:ablation_dual}
% \end{figure}

\subsection{Unified Comparison with Alignment Strategies}
We compare ReLoop with representative alignment methods, LLaVA-RLHF, HA-DPO, and POVID on both fine-grained hallucination metrics and broader benchmark evaluations. As shown in Table~\ref{tab:alignment_finegrained}, ReLoop consistently
\begin{table}[t]
\centering
\small
\setlength{\tabcolsep}{2pt}
\begin{tabular}{ccccc}
\toprule
\textbf{Method} & AMBER$\uparrow$ & MME$\uparrow$ & MMHal-B$\uparrow$ & Hallu-B$\uparrow$ \\
\midrule
LLaVA-1.5 & 73.9 & \textbf{1513} & 65.4 & 48.6 \\
+ LLaVA-RLHF               & 73.8 & 1231 & 64.3 & 43.2 \\
+ HA-DPO               & 77.2 & 1374 & 65.6 & 49.9 \\
+ POVID              & 75.8 & 1421 & 65.9 & 51.4 \\
+ \textbf{ReLoop}    & \textbf{80.3} & 1505 & \textbf{68.9} & \textbf{52.3} \\
\bottomrule
\end{tabular}
\caption{Benchmark-level comparison of ReLoop with alignment strategies across four evaluation baselines.}
\label{tab:alignment_benchmark}
\end{table}outperforms alternatives on POPE, CHAIR, F1, and faithfulness metrics, indicating stronger hallucination suppression and cross-modal faithfulness. On benchmark-level evaluations (Table~\ref{tab:alignment_benchmark}), ReLoop leads on AMBER, MMHal-B, and HallusionBench, while remaining competitive on MME. The slight MME drop may reflect a common trade-off between alignment supervision and low-level perception, also observed in other alignment-based methods like LLaVA-RLHF. These findings underscore ReLoop's effectiveness across both targeted and comprehensive settings.

% \subsection{Case Study}
% To better understand the behavior of ReLoop, we present a qualitative case study covering four representative hallucination types: object, attribute, relation, and event, summarized in Appendix~\ref{sec:appendix_case}. 
\subsection{Additional Analyses}
\label{sec:additional_analyses}
To further substantiate the effectiveness and practicality of \textsc{ReLoop}, we provide supplementary analyses in the appendices. Appendix~\ref{a6} reports a Training Cost Breakdown: cross-method cost, CFP overhead attribution, and convergence/epoch-level timing. Appendix~\ref{sec:appendix_aug_ablation} ablates contrastive augmentation to isolate closed-loop gains. Appendix~\ref{sec:appendix_case} provides a Case Study over four hallucination types and a nonsensical-answer failure mode, illustrating early rejection, ACW $\gamma$-downweighting, and entropy-aware masking.

\section{Conclusion}
We present \textbf{ReLoop}, a closed-loop training framework that mitigates hallucinations in MLLMs by enforcing semantic and visual consistency through bidirectional feedback. By incorporating language reconstruction, visual description, and attention alignment, ReLoop allows models to verify and refine predictions during training. Experiments show consistent gains in hallucination suppression and interpretability, establishing ReLoop as a general foundation for building more reliable MLLMs.

\section*{Potential Limitations}

\textbf{Performance Variability Across Hallucination Types.}
While ReLoop substantially improves hallucination suppression in object and attribute categories, its effectiveness on relation and event hallucinations remains relatively modest. These hallucination types often involve higher-order reasoning and temporal or spatial understanding, which are less easily corrected through current consistency signals. Future extensions may incorporate specialized supervision tailored to relational semantics or causal cues to address this gap.

\noindent\textbf{Supervision Dependency and Domain Adaptability.}
ReLoop relies on access to paired image–question–answer data to compute consistency signals. This requirement poses challenges in domains with limited high-quality supervision, such as medical or scientific imaging. Moreover, the training framework assumes reasonably clean and grounded reference answers, which may not hold in low-resource or noisy environments. Reducing ReLoop's dependence on strongly supervised inputs and exploring semi-supervised or synthetic feedback generation remain promising directions for broader applicability.

\noindent\textbf{Scope of evaluation.}
Our study focuses on standard VQA-style image benchmarks and general-domain LVLMs. We have not evaluated text-heavy or long-tail domains (e.g., \textit{dense OCR, charts}) or temporal reasoning tasks, where attention allocation and signal reliability may differ. Extending evaluation to these regimes is left for future work.

% Despite these limitations, we believe they also mark fruitful directions for future work. Improving coverage of complex hallucination types, reducing supervision requirements, and decoupling from fixed pretrained tools may further unlock the potential of closed-loop training for robust and interpretable MLLMs.

\section*{Ethics Statement}
All datasets utilized in this work are either publicly released or ethically sourced, ensuring full compliance with associated data usage policies. For evaluation purposes, we additionally include AI-generated content produced under controlled prompting conditions. These samples are clearly labeled and subjected to careful human verification to ensure factual accuracy and annotation quality. We acknowledge the broader implications of hallucination mitigation in AI systems and advocate for responsible model development that prioritizes reliability, fairness, and interpretability. 

% Bibliography entries for the entire Anthology, followed by custom entries
%\bibliography{anthology,custom}
% Custom bibliography entries only
\bibliography{custom}

\begin{thebibliography}{38}
\providecommand{\natexlab}[1]{#1}

\bibitem[{Alsulaimawi(2025)}]{alsulaimawi2025feedback}
Zahir Alsulaimawi. 2025.
\newblock \href {https://doi.org/10.48550/arXiv.2504.04772} {Feedback-enhanced hallucination-resistant vision-language model for real-time scene understanding}.
\newblock \emph{arXiv preprint arXiv:2504.04772}.

\bibitem[{Asai et~al.(2023)Asai, Wu, Wang, Sil, and Hajishirzi}]{asai2023selfrag}
Akari Asai, Zeqiu Wu, Yizhong Wang, Avirup Sil, and Hannaneh Hajishirzi. 2023.
\newblock \href {https://doi.org/10.48550/arXiv.2310.11511} {Self-rag: Learning to retrieve, generate, and critique through self-reflection}.
\newblock \emph{arXiv preprint arXiv:2310.11511}.

\bibitem[{Ayala and Béchard(2024)}]{ayala2024reducing}
Orlando~Marquez Ayala and Patrice Béchard. 2024.
\newblock \href {https://doi.org/10.18653/v1/2024.naacl-industry.19} {Reducing hallucination in structured outputs via retrieval-augmented generation}.
\newblock In \emph{Proceedings of the 2024 Conference of the North American Chapter of the Association for Computational Linguistics: Human Language Technologies (Volume 6: Industry Track)}, pages 228--238. Association for Computational Linguistics.

\bibitem[{Chen et~al.(2024)Chen, Zhang, Luo, Chai, and Liu}]{chen2024pad}
Ruizhe Chen, Xiaotian Zhang, Meng Luo, Wenhao Chai, and Zuozhu Liu. 2024.
\newblock Pad: Personalized alignment at decoding-time.
\newblock \emph{arXiv e-prints}, pages arXiv--2410.

\bibitem[{Fu et~al.(2023)Fu, Chen, Shen, Qin, Zhang, Lin, Yang, Zheng, Li, Sun, Wu, and Ji}]{fu2023mme}
Chaoyou Fu, Peixian Chen, Yunhang Shen, Yulei Qin, Mengdan Zhang, Xu~Lin, Jinrui Yang, Xiawu Zheng, Ke~Li, Xing Sun, Yunsheng Wu, and Rongrong Ji. 2023.
\newblock \href {https://doi.org/10.48550/arXiv.2306.13394} {Mme: A comprehensive evaluation benchmark for multimodal large language models}.
\newblock \emph{arXiv preprint arXiv:2306.13394}.

\bibitem[{Gu and Wang(2025)}]{tpo2025}
Jihao Gu and Yingyao Wang. 2025.
\newblock \href {https://doi.org/10.48550/arXiv.2412.14487} {Token preference optimization with self-calibrated visual-anchored rewards for hallucination mitigation}.
\newblock \emph{arXiv preprint arXiv:2412.14487}.

\bibitem[{Guan et~al.(2024)Guan, Liu, Wu, Xian, Li, Liu, Wang, Chen, Huang, Yacoob, Manocha, and Zhou}]{Guan_2024_CVPR}
Tianrui Guan, Fuxiao Liu, Xiyang Wu, Ruiqi Xian, Zongxia Li, Xiaoyu Liu, Xijun Wang, Lichang Chen, Furong Huang, Yaser Yacoob, Dinesh Manocha, and Tianyi Zhou. 2024.
\newblock \href {https://arxiv.org/abs/2310.14566} {Hallusionbench: An advanced diagnostic suite for entangled language hallucination and visual illusion in large vision-language models}.
\newblock In \emph{Proceedings of the IEEE/CVF Conference on Computer Vision and Pattern Recognition (CVPR)}, pages 14375--14385. IEEE.

\bibitem[{Huang et~al.(2024)Huang, Dong, Zhang, Wang, He, Wang, Lin, Zhang, and Yu}]{li2024opera}
Qidong Huang, Xiaoyi Dong, Pan Zhang, Bin Wang, Conghui He, Jiaqi Wang, Dahua Lin, Weiming Zhang, and Nenghai Yu. 2024.
\newblock \href {https://arxiv.org/abs/2311.17911} {Opera: Alleviating hallucination in multi-modal large language models via over-trust penalty and retrospection-allocation}.
\newblock In \emph{Proceedings of the IEEE/CVF Conference on Computer Vision and Pattern Recognition (CVPR)}.
\newblock Highlight Paper.

\bibitem[{Jiang et~al.(2024)Jiang, Xu, Dong, Chen, Ye, Yan, Ye, Zhang, Huang, and Zhang}]{jiang2023hallucination}
Chaoya Jiang, Haiyang Xu, Mengfan Dong, Jiaxing Chen, Wei Ye, Ming Yan, Qinghao Ye, Ji~Zhang, Fei Huang, and Shikun Zhang. 2024.
\newblock \href {https://openaccess.thecvf.com/content/CVPR2024/papers/Jiang_Hallucination_Augmented_Contrastive_Learning_for_Multimodal_Large_Language_Model_CVPR_2024_paper.pdf} {Hallucination augmented contrastive learning for multimodal large language model}.
\newblock In \emph{Proceedings of the IEEE/CVF Conference on Computer Vision and Pattern Recognition (CVPR)}, pages 27036--27045.
\newblock Code available at \url{https://github.com/X-PLUG/mPLUG-HalOwl/tree/main/hacl}.

\bibitem[{Jing et~al.(2024)Jing, Li, Chen, and Du}]{jing-etal-2024-faithscore}
Liqiang Jing, Ruosen Li, Yunmo Chen, and Xinya Du. 2024.
\newblock \href {https://doi.org/10.18653/v1/2024.findings-emnlp.290} {Faithscore: Fine-grained evaluations of hallucinations in large vision-language models}.
\newblock In \emph{Findings of the Association for Computational Linguistics: EMNLP 2024}, pages 5042--5063. Association for Computational Linguistics.

\bibitem[{Johnson-Laird(1983)}]{johnson1983mental}
Philip~N. Johnson-Laird. 1983.
\newblock \emph{Mental Models: Towards a Cognitive Science of Language, Inference, and Consciousness}.
\newblock Harvard University Press, Cambridge, MA.

\bibitem[{Kalavasis et~al.(2024)Kalavasis, Mehrotra, and Velegkas}]{kalavasis2024limits}
Alkis Kalavasis, Anay Mehrotra, and Grigoris Velegkas. 2024.
\newblock \href {https://doi.org/10.48550/arXiv.2411.09642} {On the limits of language generation: Trade-offs between hallucination and mode collapse}.
\newblock \emph{arXiv preprint arXiv:2411.09642}.

\bibitem[{Kim et~al.(2025)Kim, Jeong, Chen, Li, Lu, Alhamoud, Mun, Grau, Jung, Gameiro, Fan, Park, Lin, Yoon, Yoon, Sap, Tsvetkov, Liang, Xu, Liu, McDuff, Lee, Park, Tulebaev, and Breazeal}]{kim2025medical}
Yubin Kim, Hyewon Jeong, Shan Chen, Shuyue~Stella Li, Mingyu Lu, Kumail Alhamoud, Jimin Mun, Cristina Grau, Minseok Jung, Rodrigo Gameiro, Lizhou Fan, Eugene Park, Tristan Lin, Joonsik Yoon, Wonjin Yoon, Maarten Sap, Yulia Tsvetkov, Paul Liang, Xuhai Xu, and 6 others. 2025.
\newblock \href {https://doi.org/10.48550/arXiv.2503.05777} {Medical hallucinations in foundation models and their impact on healthcare}.
\newblock \emph{arXiv preprint arXiv:2503.05777}.

\bibitem[{Leng et~al.(2023)Leng, Zhang, Chen, Li, Lu, Miao, and Bing}]{damonlpsg2023vcd}
Sicong Leng, Hang Zhang, Guanzheng Chen, Xin Li, Shijian Lu, Chunyan Miao, and Lidong Bing. 2023.
\newblock \href {https://arxiv.org/abs/2311.16922} {Mitigating object hallucinations in large vision-language models through visual contrastive decoding}.
\newblock \emph{arXiv preprint arXiv:2311.16922}.

\bibitem[{Leng et~al.(2024)Leng, Zhang, Chen, Li, Lu, Miao, and Bing}]{leng2023mitigating}
Sicong Leng, Hang Zhang, Guanzheng Chen, Xin Li, Shijian Lu, Chunyan Miao, and Lidong Bing. 2024.
\newblock \href {https://openaccess.thecvf.com/content/CVPR2024/papers/Leng_Mitigating_Object_Hallucinations_in_Large_Vision-Language_Models_through_Visual_Contrastive_CVPR_2024_paper.pdf} {Mitigating object hallucinations in large vision-language models through visual contrastive decoding}.
\newblock In \emph{Proceedings of the IEEE/CVF Conference on Computer Vision and Pattern Recognition (CVPR)}, pages 1316--1325.
\newblock Poster Highlight.

\bibitem[{Li et~al.(2024)Li, Lu, Fei, Luo, Dai, Xia, Jin, Gan, Qi, Fu et~al.}]{li2024survey}
Jian Li, Weiheng Lu, Hao Fei, Meng Luo, Ming Dai, Min Xia, Yizhang Jin, Zhenye Gan, Ding Qi, Chaoyou Fu, and 1 others. 2024.
\newblock A survey on benchmarks of multimodal large language models.
\newblock \emph{arXiv preprint arXiv:2408.08632}.

\bibitem[{Li et~al.(2023{\natexlab{a}})Li, Li, Savarese, and Hoi}]{li2023blip2}
Junnan Li, Dongxu Li, Silvio Savarese, and Steven Hoi. 2023{\natexlab{a}}.
\newblock \href {https://doi.org/10.48550/arXiv.2301.12597} {Blip-2: Bootstrapping language-image pre-training with frozen image encoders and large language models}.
\newblock \emph{arXiv preprint arXiv:2301.12597}.

\bibitem[{Li et~al.(2025{\natexlab{a}})Li, Cao, He, Cheng, Fu, Xiao, Wang, and Tang}]{li2025miv}
Yanshu Li, Yi~Cao, Hongyang He, Qisen Cheng, Xiang Fu, Xi~Xiao, Tianyang Wang, and Ruixiang Tang. 2025{\natexlab{a}}.
\newblock \href {https://openreview.net/forum?id=9ffYcEiNw9} {M{\texttwosuperior}{IV}: Towards efficient and fine-grained multimodal in-context learning via representation engineering}.
\newblock In \emph{Second Conference on Language Modeling}.

\bibitem[{Li et~al.(2025{\natexlab{b}})Li, Yun, Yang, Feng, Huang, and Tang}]{li2025taco}
Yanshu Li, Tian Yun, Jianjiang Yang, Pinyuan Feng, Jinfa Huang, and Ruixiang Tang. 2025{\natexlab{b}}.
\newblock Taco: Enhancing multimodal in-context learning via task mapping-guided sequence configuration.
\newblock \emph{arXiv preprint arXiv:2505.17098}.

\bibitem[{Li et~al.(2023{\natexlab{b}})Li, Du, Zhou, Wang, Zhao, and Wen}]{Li-hallucination-2023}
Yifan Li, Yifan Du, Kun Zhou, Jinpeng Wang, Wayne~Xin Zhao, and Ji-Rong Wen. 2023{\natexlab{b}}.
\newblock \href {https://openreview.net/forum?id=xozJw0kZXF} {Evaluating object hallucination in large vision-language models}.
\newblock In \emph{Proceedings of the 2023 Conference on Empirical Methods in Natural Language Processing}.

\bibitem[{Liu et~al.(2023{\natexlab{a}})Liu, Li, Li, and Lee}]{liu2023improvedllava}
Haotian Liu, Chunyuan Li, Yuheng Li, and Yong~Jae Lee. 2023{\natexlab{a}}.
\newblock \href {https://doi.org/10.48550/arXiv.2310.03744} {Improved baselines with visual instruction tuning}.
\newblock \emph{arXiv preprint arXiv:2310.03744}.

\bibitem[{Liu et~al.(2023{\natexlab{b}})Liu, Li, Wu, and Lee}]{liu2023visual}
Haotian Liu, Chunyuan Li, Qingyang Wu, and Yong~Jae Lee. 2023{\natexlab{b}}.
\newblock \href {https://doi.org/10.48550/arXiv.2304.08485} {Visual instruction tuning}.
\newblock \emph{arXiv preprint arXiv:2304.08485}.

\bibitem[{Luo et~al.(2024{\natexlab{a}})Luo, Fei, Li, Wu, Liu, Poria, Cambria, Lee, and Hsu}]{luo2024panosent}
Meng Luo, Hao Fei, Bobo Li, Shengqiong Wu, Qian Liu, Soujanya Poria, Erik Cambria, Mong-Li Lee, and Wynne Hsu. 2024{\natexlab{a}}.
\newblock Panosent: A panoptic sextuple extraction benchmark for multimodal conversational aspect-based sentiment analysis.
\newblock In \emph{Proceedings of the 32nd ACM International Conference on Multimedia}, pages 7667--7676.

\bibitem[{Luo et~al.(2025)Luo, Wu, Jing, Ju, Zheng, Lai, Wu, Du, Li, Yan, Luo, Wang, Fei, Lee, and Hsu}]{luo2025drvhierarchicalperceptiontemporalcognitionframework}
Meng Luo, Shengqiong Wu, Liqiang Jing, Tianjie Ju, Li~Zheng, Jinxiang Lai, Tianlong Wu, Xinya Du, Jian Li, Siyuan Yan, Jiebo Luo, William~Yang Wang, Hao Fei, Mong-Li Lee, and Wynne Hsu. 2025.
\newblock \href {https://arxiv.org/abs/2509.11866} {Dr.v: A hierarchical perception-temporal-cognition framework to diagnose video hallucination by fine-grained spatial-temporal grounding}.
\newblock \emph{Preprint}, arXiv:2509.11866.

\bibitem[{Luo et~al.(2024{\natexlab{b}})Luo, Zhang, Wu, Li, Han, and Fei}]{luo2024nus}
Meng Luo, Han Zhang, Shengqiong Wu, Bobo Li, Hong Han, and Hao Fei. 2024{\natexlab{b}}.
\newblock Nus-emo at semeval-2024 task 3: Instruction-tuning llm for multimodal emotion-cause analysis in conversations.
\newblock \emph{arXiv preprint arXiv:2501.17261}.

\bibitem[{Mala et~al.(2025)Mala, Gezici, and Giannotti}]{mala2025hybrid}
Chandana~Sree Mala, Gizem Gezici, and Fosca Giannotti. 2025.
\newblock \href {https://doi.org/10.48550/arXiv.2504.05324} {Hybrid retrieval for hallucination mitigation in large language models: A comparative analysis}.
\newblock \emph{arXiv preprint arXiv:2504.05324}.

\bibitem[{{OpenAI}(2023)}]{openai2023gpt4v}
{OpenAI}. 2023.
\newblock \href {https://doi.org/10.48550/arXiv.2303.08774} {Gpt-4 technical report}.
\newblock \emph{arXiv preprint arXiv:2303.08774}.

\bibitem[{Park et~al.(2023)Park, Xiao, Warnell, Yedidsion, and Stone}]{park2023learning}
Jin-Soo Park, Xuesu Xiao, Garrett Warnell, Harel Yedidsion, and Peter Stone. 2023.
\newblock \href {https://doi.org/10.1109/ICRA48891.2023.10161327} {Learning perceptual hallucination for multi-robot navigation in narrow hallways}.
\newblock In \emph{Proceedings of the 2023 IEEE International Conference on Robotics and Automation (ICRA)}, London, England.

\bibitem[{Rohrbach et~al.(2018)Rohrbach, Hendricks, Burns, Darrell, and Saenko}]{rohrbach2018object}
Anna Rohrbach, Lisa~Anne Hendricks, Kaylee Burns, Trevor Darrell, and Kate Saenko. 2018.
\newblock \href {https://aclanthology.org/D18-1437/} {Object hallucination in image captioning}.
\newblock In \emph{Proceedings of the 2018 Conference on Empirical Methods in Natural Language Processing}, pages 4035--4045. Association for Computational Linguistics.

\bibitem[{Sennrich et~al.(2016)Sennrich, Haddow, and Birch}]{sennrich2016edinburgh}
Rico Sennrich, Barry Haddow, and Alexandra Birch. 2016.
\newblock \href {https://doi.org/10.18653/v1/W16-2323} {Edinburgh neural machine translation systems for wmt 16}.
\newblock In \emph{Proceedings of the First Conference on Machine Translation: Volume 2, Shared Task Papers}, pages 371--376, Berlin, Germany. Association for Computational Linguistics.

\bibitem[{Shao et~al.(2024)Shao, Wang, Zhu, Xu, Song, Bi, Zhang, Zhang, Li, Wu, and Guo}]{deepseek2024grpo}
Zhihong Shao, Peiyi Wang, Qihao Zhu, Runxin Xu, Junxiao Song, Xiao Bi, Haowei Zhang, Mingchuan Zhang, Y.~K. Li, Y.~Wu, and Daya Guo. 2024.
\newblock \href {https://doi.org/10.48550/arXiv.2402.03300} {Deepseekmath: Pushing the limits of mathematical reasoning in open language models}.
\newblock \emph{arXiv preprint arXiv:2402.03300}.

\bibitem[{Sun et~al.(2023)Sun, Shen, Cao, Liu, Li, Shen, Gan, Gui, Wang, Yang, Keutzer, and Darrell}]{sun2023aligning}
Zhiqing Sun, Sheng Shen, Shengcao Cao, Haotian Liu, Chunyuan Li, Yikang Shen, Chuang Gan, Liang-Yan Gui, Yu-Xiong Wang, Yiming Yang, Kurt Keutzer, and Trevor Darrell. 2023.
\newblock \href {https://doi.org/10.48550/arXiv.2309.14525} {Aligning large multimodal models with factually augmented rlhf}.
\newblock \emph{arXiv preprint arXiv:2309.14525}.

\bibitem[{Sun et~al.(2024)Sun, Zang, Zheng, Song, Xu, Zhang, Yu, Song, and Li}]{sun2024redeep}
Zhongxiang Sun, Xiaoxue Zang, Kai Zheng, Yang Song, Jun Xu, Xiao Zhang, Weijie Yu, Yang Song, and Han Li. 2024.
\newblock \href {https://doi.org/10.48550/arXiv.2410.11414} {Redeep: Detecting hallucination in retrieval-augmented generation via mechanistic interpretability}.
\newblock \emph{arXiv preprint arXiv:2410.11414}.

\bibitem[{Villa et~al.(2025)Villa, Alcázar, Alfarra, Araujo, Soto, and Ghanem}]{villa2025eagle}
Andrés Villa, Juan~León Alcázar, Motasem Alfarra, Vladimir Araujo, Alvaro Soto, and Bernard Ghanem. 2025.
\newblock \href {https://doi.org/10.48550/arXiv.2501.02699} {Eagle: Enhanced visual grounding minimizes hallucinations in instructional multimodal models}.
\newblock \emph{arXiv preprint arXiv:2501.02699}.

\bibitem[{Wang et~al.(2023)Wang, Wang, Xu, Zhang, Gu, Jia, Wang, Xu, Yan, Zhang, and Sang}]{wang2023llm}
Junyang Wang, Yuhang Wang, Guohai Xu, Jing Zhang, Yukai Gu, Haitao Jia, Jiaqi Wang, Haiyang Xu, Ming Yan, Ji~Zhang, and Jitao Sang. 2023.
\newblock \href {https://doi.org/10.48550/arXiv.2311.07397} {An llm-free multi-dimensional benchmark for mllms hallucination evaluation}.
\newblock \emph{arXiv preprint arXiv:2311.07397}.

\bibitem[{Yin et~al.(2023)Yin, Fu, Zhao, Xu, Wang, Sui, Shen, Li, Sun, and Chen}]{woodpecker2023}
Shukang Yin, Chaoyou Fu, Sirui Zhao, Tong Xu, Hao Wang, Dianbo Sui, Yunhang Shen, Ke~Li, Xing Sun, and Enhong Chen. 2023.
\newblock \href {https://doi.org/10.48550/arXiv.2310.16045} {Woodpecker: Hallucination correction for multimodal large language models}.
\newblock \emph{arXiv preprint arXiv:2310.16045}.

\bibitem[{Zhao et~al.(2023)Zhao, Wang, Ouyang, Dong, Wang, and He}]{zhao2023beyond}
Zhiyuan Zhao, Bin Wang, Linke Ouyang, Xiaoyi Dong, Jiaqi Wang, and Conghui He. 2023.
\newblock \href {https://doi.org/10.48550/arXiv.2311.16839} {Beyond hallucinations: Enhancing lvlms through hallucination-aware direct preference optimization}.
\newblock \emph{arXiv preprint arXiv:2311.16839}.

\bibitem[{Zhou et~al.(2024)Zhou, Cui, Rafailov, Finn, and Yao}]{zhou2024povid}
Yiyang Zhou, Chenhang Cui, Rafael Rafailov, Chelsea Finn, and Huaxiu Yao. 2024.
\newblock \href {https://doi.org/10.48550/arXiv.2402.11411} {Aligning modalities in vision large language models via preference fine-tuning}.
\newblock \emph{arXiv preprint arXiv:2402.11411}.

\end{thebibliography}

\newpage
\appendix
\section{Additional Experimental Details}
\subsection{Implementation Details}
\label{a1}
\paragraph{Backbone and Setup.}
We apply ReLoop to five representative LVLMs with diverse architectures: MiniGPT-4, InstructBLIP, LLaVA-1.5, mPLUG-owl, and ShareGPT4V. To assess generalizability on stronger backbones, we further evaluate on LLaVA-1.6 and Qwen-VL-2.5 (Table~\ref{tab:model_performance}). Importantly, we do not alter the internal structures of these models. ReLoop is introduced as a lightweight, external consistency-supervision framework during training. All backbones are initialized with their public checkpoints and keep their visual encoders (e.g., \textit{ViT, CLIP}) frozen.

\paragraph{ReLoop Components.}
ReLoop introduces three frozen feedback modules:  
(1) CFP-Lang (MiniGPT-4-based reverse question reconstructor);  
(2) CFP-Vis (BLIP-2-based visual describer);  
(3) Attention Supervision that aligns decoder attention maps with entropy-based soft pseudo-labels.  
A frozen BERT encoder plus an MLP scorer serves as a lightweight semantic aggregator. All feedback modules remain frozen; only the backbone and the aggregator are updated.

\paragraph{Training Details.}
Experiments are performed on 8$\times$A100 GPUs (80GB) using mixed-precision training (fp16) for 3 epochs. We adopt the AdamW optimizer with parameters $\beta_1 = 0.9$, $\beta_2 = 0.98$, and a weight decay of 0.05. The effective batch size is 128, with a gradient accumulation step of 8. The initial learning rate is set to $5 \times 10^{-5}$, along with 1{,}000 warm-up steps and cosine learning rate decay scheduling. Unless otherwise stated, all accuracy and resource measurements follow this main setup. For efficiency-only timing and method-level comparability, we additionally report a controlled regime: 4$\times$A100 GPUs, batch size $12$/GPU, and a fixed 2k-step schedule.

\paragraph{Loss Function.}
The overall objective is
\begin{equation}
\mathcal{L}_{\text{total}} = \mathcal{L}_{\text{sft}}
+ \alpha\,\mathcal{L}_{\text{align}}
+ \beta\,\mathcal{L}_{\text{vis}}
+ \gamma\,\mathcal{L}_{\text{attn}}
+ \lambda\,\Omega(\theta)
\end{equation}
We set the hyper-parameters as $\alpha = 1.0$, $\beta = 0.7$, and $\lambda = 10^{-5}$. The weight $\gamma$ is dynamically adjusted by the Adaptive Consistency Weighting (ACW) mechanism, which modulates $\gamma$ based on the BERTScore between the original and reconstructed questions (see Section~\ref{sec:acw}).

\subsection{Training Dataset Construction}
\label{sec:appendix_dataset}
We curated approximately 30K high-quality QA-image triplets from the LLaVA-Instruct-150K corpus \citep{liu2023improvedllava}, each containing an image, an open-ended question, and a human-annotated answer. To simulate hallucination supervision, we generated semantically contradictory answers by modifying key elements (e.g., \textit{objects, attributes, or relations}) in the references. These hallucinated samples were automatically constructed and manually verified for quality and type diversity. In Experiment~\ref{6.1}, we selected 500 representative QA-image pairs from the filtered validation set based on POPE and MMHalBench, equally split between hallucinated and non-hallucinated cases. In Experiment~\ref{6.2}, we evaluated five LVLMs on this curated set to assess the impact of ReLoop. Models with open alignment architectures (e.g., \textit{MiniGPT-4, InstructBLIP}) showed the greatest improvement, while high-performing black-box models (e.g., \textit{ShareGPT4V}) saw minimal gains, suggesting ReLoop's effectiveness hinges on alignment signal compatibility.

\subsection{Evaluation Metrics}
\label{sec:appendix_metrics}
To comprehensively evaluate the effectiveness of ReLoop in mitigating hallucinations and enhancing visual grounding, we adopt a structured set of metrics covering both hallucination suppression and cross-modal consistency. In particular, shown in Table~\ref{tab:alignment_finegrained}, we group the metrics into two key categories: \textit{Hallucination Suppression}, which quantifies the presence of non-existent or spurious content, and \textit{Cross-modal Faithfulness}, which assesses the semantic and perceptual alignment between generated text and visual input.
\subsubsection{Metrics on Hallucination Suppression}
For hallucination evaluation, we incorporate CHAIR~\cite{rohrbach2018object} to measure hallucination frequencies at instance levels and include POPE~\cite{Li-hallucination-2023}, a probing-based diagnostic benchmark to evaluate object hallucinations through direct VQA-style interactions. Together, these metrics allow us to holistically assess ReLoop's ability to suppress hallucinated content while preserving descriptive quality.
\begin{itemize}[leftmargin=*]
\item \textbf{CHAIR}~\cite{rohrbach2018object} (Caption Hallucination Assessment with Image Relevance) quantifies hallucinations by detecting whether the model-generated captions mention objects that do not exist in the image. It provides two variants:
\begin{equation}
\text{CHAIR}_I = \frac{|\{\text{hallucinated objects}\}|}{|\{\text{all objects}\}|}
\end{equation}
\begin{equation}
\text{CHAIR}_S = \frac{|\{\text{hallucinated responses}\}|}{|\{\text{all responses}\}|}
\end{equation}

where CHAIR$_I$ measures instance-level hallucination (object granularity) and CHAIR$_S$ measures sentence-level hallucination (response granularity).

\item \textbf{POPE}~\cite{Li-hallucination-2023} (Polling-based Object Probing Evaluation) automates hallucination detection via instance-level object probing. It:
\begin{itemize}
\item Segments objects in the image;
\item Asks the model about object existence and introduces distractor queries;
\item Computes metrics such as F1 score to measure detection precision.
\end{itemize}
POPE offers direct insights into a model's visual grounding capability through objective visual questioning.
\end{itemize}
\subsubsection{Metrics on Cross-modal Faithfulness}
On the side of Cross-modal Faithfulness, we adopt Faith and Faith$_S$~\citep{jing-etal-2024-faithscore}, which evaluate how well the generated text is grounded in the visual input. Faith focuses on overall alignment, while Faith$_S$ specifically checks whether statements are supported by the visual evidence in a token-level or segment-wise manner. In addition, we report the $F1$ score, a standard metric that captures the harmonic mean of precision and recall between the predicted and reference entities. In our context, it reflects how well the model identifies relevant visual content without fabricating or omitting essential elements, thus serving as a practical indicator of the model's grounding precision and completeness.
\begin{itemize}[leftmargin=*]
\item \textbf{F1 Score} reflects the harmonic mean of precision and recall in detecting whether queried objects exist. High F1 indicates accurate recognition and rejection of hallucinated entities:
\begin{equation}
\text{F1} = 2 \cdot \frac{\text{Precision} \cdot \text{Recall}}{\text{Precision} + \text{Recall}}
\end{equation}

\item \textbf{Faith}~\citep{jing-etal-2024-faithscore} measures the overall semantic alignment between image and response. It uses automated matching or human verification to assess whether the content is factually grounded in the image:
\begin{equation}
\text{Faith} = \frac{|\text{Aligned Statements}|}{|\text{Total Statements}|}
\end{equation}

\item \textbf{Faith$_S$}~\citep{jing-etal-2024-faithscore} extends Faith to a finer granularity by evaluating the support of specific sentence segments or tokens using cross-modal supervision or saliency alignment:
\begin{equation}
\text{Faith}_S = \frac{|\text{Grounded Segments or Tokens}|}{|\text{Total Segments or Tokens}|}
\end{equation}

\end{itemize}

\subsection{Evaluation Benchmark}
\label{a4}
Besides, to provide a fine-grained and multi-perspective assessment of ReLoop's effectiveness in suppressing hallucinations and enhancing cross-modal faithfulness, we adopt four complementary benchmarks. AMBER~\citep{wang2023llm} targets object-level hallucinations, while MMHal-B~\citep{sun2023aligning} and HallusionBench~\citep{Guan_2024_CVPR} assess errors in attributes, spatial relations, and perceptual consistency. MME~\citep{fu2023mme} covers general multimodal capabilities such as OCR and counting. These benchmarks collectively evaluate generative and discriminative capabilities, entity grounding, perceptual consistency, and multimodal reasoning:

\begin{itemize}[leftmargin=1em]
    \item \textbf{AMBER}~\citep{wang2023llm}: An LLM-free multi-dimensional benchmark that diagnoses hallucinations in both generative and discriminative tasks. It explicitly tests object \textit{existence}, \textit{attributes}, and \textit{relations}, allowing us to assess ReLoop's object-level grounding fidelity, attribute correctness, and relational accuracy. This supports the evaluation of semantic precision in visual grounding.

    \item \textbf{MMHal-B}~\citep{sun2023aligning}: A benchmark built upon fact-augmented reinforcement learning (RLHF) that penalizes hallucinated attributes and spatial configurations. MMHal-B offers targeted diagnostics for hallucination suppression in factual and compositional dimensions, particularly assessing whether ReLoop can resist overgeneralization and maintain factual grounding under complex prompts.

    \item \textbf{HallusionBench}~\citep{Guan_2024_CVPR}: A benchmark that probes visual-linguistic robustness under ambiguous image-text settings. It emphasizes contextual grounding, requiring models to handle subtle visual cues and nuanced linguistic traps. HallusionBench evaluates ReLoop's ability to maintain perceptual consistency and reject misleading contextual cues that typically trigger hallucinations.

    \item \textbf{MME}~\cite{fu2023mme}: A broad-spectrum benchmark measuring multimodal perception and cognition across 14 sub-tasks, including OCR, object counting, spatial reasoning, and commonsense grounding. MME validates whether ReLoop's structured supervision translates into generalized improvements in visual understanding and multimodal reasoning, beyond hallucination mitigation.
\end{itemize}

Together, these benchmarks offer layered supervision signals from fine-grained object hallucination detection to holistic multimodal cognition, providing strong empirical evidence of ReLoop's reliability across diverse real-world tasks.
\begin{figure*}[t]
    \centering
    \includegraphics[width=\textwidth]{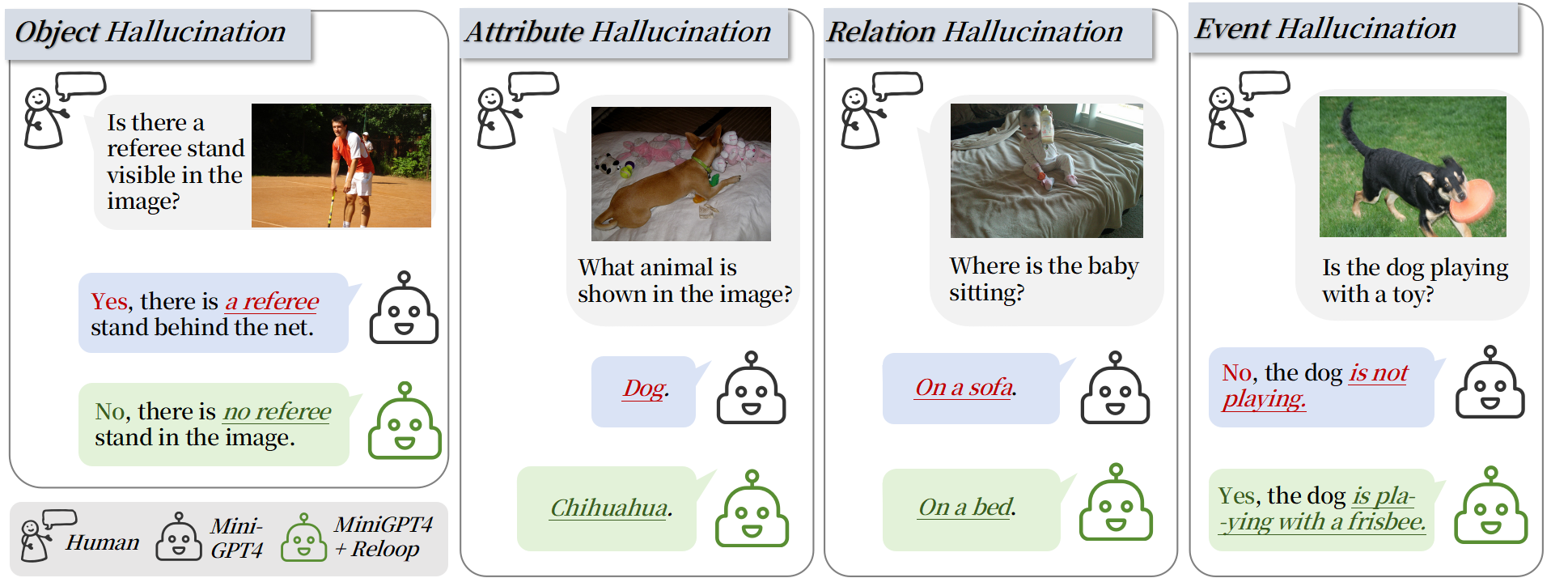}
    \caption{
    \textbf{Case Study:} Comparison between MiniGPT-4 and ReLoop across four types of hallucination in open-ended VQA: \textit{Object}, \textit{Attribute}, \textit{Relation}, and \textit{Event}. ReLoop produces more accurate and grounded responses by aligning its outputs with both the visual evidence and the question semantics.
    }
    \label{fig:case_study}
\end{figure*}
\subsection{Baseline Implementation}
\label{sec:appendix_alignment}

To evaluate ReLoop's generalizability and additive benefit, we compare it with three representative alignment-based hallucination mitigation strategies: LLaVA-RLHF~\cite{sun2023aligning}, HA-DPO~\cite{zhao2023beyond}
, and POVID~\cite{zhou2024povid}. These baselines span a diverse range of supervision paradigms, from reinforcement learning to contrastive grounding. Importantly, all methods are applied on top of the same backbone (LLaVA-1.5) with consistent training configurations, ensuring fair comparison.
\begin{itemize}[leftmargin=1em]
\item \textbf{LLaVA-RLHF}~\cite{sun2023aligning} aligns responses to human preferences through reinforcement learning from human feedback. While effective for improving general fluency and tone, it does not explicitly penalize visual or factual inconsistencies.
\item \textbf{HA-DPO}~\cite{zhao2023beyond}
 adopts hallucination-aware preference optimization by contrasting faithful versus hallucinated generations. This method introduces targeted loss signals during fine-tuning, encouraging the model to avoid semantically spurious content.
\item \textbf{POVID}~\cite{zhou2024povid} enhances visual grounding via perturbed image inputs, injecting contrastive visual signals to reduce reliance on textual priors and promote visual fidelity.
\item \textbf{VCD}~\cite{damonlpsg2023vcd} is a training-free, inference-time method that contrasts output distributions from the original image and a visually distorted counterpart, down-weighting tokens supported mainly by unimodal/language priors and thereby mitigating object hallucinations without further tuning.
\end{itemize}
Results from both fine-grained hallucination metrics (Table~\ref{tab:alignment_finegrained}) and benchmark-level evaluations (Table~\ref{tab:alignment_benchmark}) demonstrate that ReLoop consistently outperforms all competing methods. These results validate ReLoop as a robust and generalizable framework capable of enhancing multimodal model performance beyond what is achievable by current alignment-based techniques alone.
\begin{table*}[t]
\centering
\renewcommand{\arraystretch}{0.9}
\small
\setlength{\tabcolsep}{1.5pt}

% 定义便捷的列类型：L/C 基于 m{}，可实现单元格内容垂直居中
\newcolumntype{L}[1]{>{\raggedright\arraybackslash}m{#1}}
\newcolumntype{C}[1]{>{\centering\arraybackslash}m{#1}}

\begin{tabular}{
  L{2.2cm}  %C Method
  C{2.1cm}  % Trainable Params
  C{1.8cm}  % Feedback Modules
  C{1.4cm}  % GPU Hours
  C{1.6cm}  % GPU Type
  C{1.6cm}  % Peak VRAM
  L{2.4cm}  % Notes
}
\toprule
\textbf{Method} &
\shortstack{\textbf{Trainable}\\\textbf{Params}} &
\shortstack{\textbf{Feedback}\\\textbf{Modules}} &
\shortstack{\textbf{GPU}\\\textbf{Hours}} &
\shortstack{\textbf{GPU}\\\textbf{Type}} &
\shortstack{\textbf{Peak}\\\textbf{VRAM}} &
\textbf{Notes} \\
\midrule
\shortstack[l]{ReLoop\\(MiniGPT-4)} &
$M{+}S$ ($\sim$1.2B) &
Frozen &
3 &
A100-40G &
$\sim$26GB &
\shortstack[l]{Fastest;\\ablation} \\
\addlinespace[2pt]
\shortstack[l]{ReLoop\\(LLaVA-1.5)} &
$M{+}S$ ($\sim$13B) &
Frozen &
6 &
A100-80G &
$\sim$48GB &
\shortstack[l]{Core experiment.\\(Sec.~5.2)} \\
\addlinespace[2pt]
\shortstack[l]{RLHF\\finetuning} &
\shortstack{$\,$Whole model\\($\sim$13B)} &
End-to-end &
20 &
A100-80G &
$\sim$70GB &
-- \\
\addlinespace[2pt]
DPO &
Whole model &
End-to-end &
8--16 &
A100-80G &
$>$50GB &
-- \\
\addlinespace[2pt]
\shortstack[l]{Contrastive\\alignment} &
Whole model &
\shortstack{Encoder\\fusion} &
$\sim$10 &
A100-80G &
$\sim$32GB &
-- \\
\bottomrule
\end{tabular}
\caption{Cross-method training cost. ReLoop updates only $M$ and $S$ with frozen CFPs, reducing optimizer/memory footprint relative to end-to-end baselines.}
\label{tab:eff_cross_method}
\end{table*}

\begin{table*}[t]
\centering
\renewcommand{\arraystretch}{0.95}
\small
\setlength{\tabcolsep}{2pt}
\begin{tabular}{
  >{\raggedright\arraybackslash}p{3.0cm}
  >{\centering\arraybackslash}p{3.0cm}
  >{\centering\arraybackslash}p{2.0cm}
  >{\centering\arraybackslash}p{2.0cm}
  >{\centering\arraybackslash}p{2.5cm}
  >{\centering\arraybackslash}p{1.7cm}
}
\toprule
\textbf{Variant} & \textbf{CFP Modules} & \textbf{GPU Hours} & \textbf{Peak Mem} & \textbf{Train Latency} & \textbf{POPE} $\uparrow$ \\
\midrule
ReLoop (Full)     & Lang + Vis   & 6.0  & 48GB & 1.8$\times$ & 82.2 \\
w/o Lang CFP      & Vis only     & 5.2  & 40GB & 1.4$\times$ & 81.5 \\
w/o Vis CFP       & Lang only    & 4.7  & 38GB & 1.3$\times$ & 81.1 \\
No CFP            & None         & 3.0  & 26GB & 1.0$\times$ & 77.2 \\
\bottomrule
\end{tabular}
\caption{CFP-component ablations (LLaVA-1.5). Each CFP adds $\sim$1 GPU hour and $\sim$10--12GB VRAM, yielding $+$3--5 POPE. CFPs are frozen yet incur forward attention/encoding; batch size may reduce (e.g., \textit{16$\rightarrow$12 on A100}).}
\label{tab:eff_cfp_ablation}
\end{table*}
\subsection{Training Cost Breakdown}
\label{a6}

We quantify ReLoop's computational footprint and efficiency trade-offs from four angles: (i) cross-method training cost, (ii) overhead attribution via CFP-component ablations, (iii) convergence versus training steps, and (iv) per-epoch cost under a controlled regime. Unless otherwise stated, CFP modules are frozen (no gradient), and only the main model $M$ and the Semantic Aggregator $S$ are updated.

\subsubsection{Cross-Method Training Cost}
\label{a6_cross}
\noindent We compare ReLoop against representative alignment methods. As summarized in Table~\ref{tab:eff_cross_method}, ReLoop updates only $M$ and $S$ with frozen CFPs, substantially reducing optimizer/memory footprint compared to end-to-end baselines. Despite extra forward passes for CFPs, ReLoop avoids optimizer states for large frozen modules, yielding substantially lower GPU hours and memory than end-to-end RLHF/DPO. This design preserves training affordability while enabling closed-loop supervision.

\subsubsection{Overhead Attribution via CFP Components}
\label{a7_cfp}
\noindent We attribute runtime and memory overheads to individual CFP branches on LLaVA-1.5, keeping data/backbone constant. Table~\ref{tab:eff_cfp_ablation} reports GPU hours, memory, latency, and POPE. As seen in Table~\ref{tab:eff_cfp_ablation}, both language and visual CFPs contribute positively to hallucination suppression, with additive gains. The per-branch overhead is bounded and predictable. Since CFPs are optional at inference, the training-time penalty does not translate to deployment latency. Both language and visual CFPs contribute positively to hallucination suppression, with additive gains. The overhead is bounded and predictable per component. Since CFPs are optional at inference, the training-time penalty does not translate to deployment latency.

\subsubsection{Convergence and Epoch-Level Cost}
\label{a8_steps_epoch}
\noindent We study returns versus training steps and report per-epoch cost under the controlled regime (4$\times$A100, batch $12$/GPU, fixed 2k steps) for method-level comparability. Step-wise gains are summarized in Table~\ref{tab:eff_steps}; the end-to-end throughput and slowdown are reported in Table~\ref{tab:eff_epoch}.\footnote{Scores here follow the controlled 2k-step regime and are not directly comparable to full-training results elsewhere.} Table~\ref{tab:eff_steps} shows fast early gains with diminishing returns beyond 2k steps. Under a fixed budget, Table~\ref{tab:eff_epoch} indicates that a moderate training slow-down trades for sizable hallucination reduction, while inference cost remains unchanged by dropping CFPs at test time.

\begin{table*}[t]
\centering
\renewcommand{\arraystretch}{0.9}
\small
\setlength{\tabcolsep}{1.5pt}

% 垂直居中列类型：左对齐 L{} / 居中 C{}
\newcolumntype{L}[1]{>{\raggedright\arraybackslash}m{#1}}
\newcolumntype{C}[1]{>{\centering\arraybackslash}m{#1}}

\begin{tabular}{
  L{2.8cm}  % Model
  C{1.5cm}  % Time/Epoch
  C{2.0cm}  % Total GPU Hrs (2k)
  C{1.3cm}  % Peak Mem
  C{2.3cm}  % Throughput
  C{1.8cm}  % Slowdown vs. Base
  C{1.3cm}  % POPE
  C{1.5cm}  % CFP @ Inference
}
\toprule
\textbf{Model} &
\shortstack{\textbf{Time}\\\textbf{/ Epoch}} &
\shortstack{\textbf{Total GPU}\\\textbf{Hrs (2k)}} &
\shortstack{\textbf{Peak}\\\textbf{Memory}} &
\shortstack{\textbf{Throughput}\\\textbf{(img/s/GPU)}} &
\shortstack{\textbf{Slowdown}\\\textbf{vs.\ Base}} &
\textbf{POPE}$\uparrow$ &
\shortstack{\textbf{CFP @}\\\textbf{Inference}} \\
\midrule
LLaVA-1.5 (base) &
\shortstack{$\sim$58\\min} &
3.0 &
26GB &
11.6 &
1.0$\times$ &
77.2 &
N/A \\
\addlinespace[2pt]
\shortstack[l]{LLaVA-1.5\\+ ReLoop} &
\shortstack{$\sim$103\\min} &
6.0 &
48GB &
6.4 &
1.77$\times$ &
82.2 &
\shortstack{Optional\\(off)} \\
\bottomrule
\end{tabular}
\caption{Per-epoch cost under the controlled regime. ReLoop increases epoch time by $\sim$77\% yet converges within comparable steps. CFPs are disabled at inference, so deployment latency remains unchanged.}
\label{tab:eff_epoch}
\end{table*}
\begin{table}[t]
\centering
\renewcommand{\arraystretch}{0.95}
\small
\setlength{\tabcolsep}{2pt}
\newcolumntype{L}[1]{>{\raggedright\arraybackslash}m{#1}}
\newcolumntype{C}[1]{>{\centering\arraybackslash}m{#1}}
\begin{tabular}{C{1.7cm} C{1.35cm} C{1.1cm} C{1.25cm} C{1.25cm}}
\toprule
\textbf{Steps} &
\textbf{GPU hours} &
\textbf{POPE}$\uparrow$ &
$\boldsymbol{\Delta}$\textbf{POPE} &
\shortstack{\textbf{Gain}\\\textbf{/ hour}} \\
\midrule
\shortstack{0\\(no Training)} & 0.0  & 77.2 & --   & --   \\
1k                      & 3.6  & 80.1 & +2.9 & 0.81 \\
2k                      & 7.2  & 81.6 & +1.5 & 0.21 \\
3k                      & 10.9 & 82.2 & +0.6 & 0.08 \\
\bottomrule
\end{tabular}
\caption{Diminishing returns with steps (LLaVA-1.5). $>$90\% of gains appear within 2k steps ($\sim$7 GPU hours).}
\label{tab:eff_steps}
\end{table}

\subsection{Contrastive Augmentation Ablation}
\label{sec:appendix_aug_ablation}

To evaluate whether \textsc{ReLoop} depends on the manually perturbed semantic negatives (Appendix~\ref{sec:appendix_dataset}), we ablate this contrastive augmentation and train on the standard LLaVA-Instruct data only, keeping all other settings identical to the main experiments. Even without contrastive augmentation, \textsc{ReLoop} significantly improves over the LLaVA-1.5 baseline on all metrics (Table~\ref{tab:abl_aug_app}), demonstrating that the core closed-loop alignment mechanism is effective when trained purely on standard supervision data. The contrastive examples provide further refinement, most notably on AMBER and MMHal-B, but they are not essential for \textsc{ReLoop} to outperform existing alignment strategies.

\begin{table}[t]
\centering
\small
\setlength{\tabcolsep}{2pt}
\renewcommand{\arraystretch}{1.05}
\newcolumntype{L}[1]{>{\raggedright\arraybackslash}m{#1}}
\newcolumntype{C}[1]{>{\centering\arraybackslash}m{#1}}
\begin{tabular}{C{1.5cm}cccc}
\toprule
\textbf{Method} & \textbf{AMBER}$\uparrow$ & \textbf{MME}$\uparrow$ & \textbf{MMHal\text{-}B}$\uparrow$ & \textbf{Hallu\text{-}B}$\uparrow$ \\
\midrule
LLaVA-1.5\\(baseline)            & 73.9 & \textbf{1513} & 65.4 & 48.6 \\
ReLoop\\(w/o augmentation)       & 78.5 & 1452          & 67.1 & 50.1 \\
\textbf{Full}\\\textbf{ReLoop}   & \textbf{80.3} & 1505 & \textbf{68.9} & \textbf{52.3} \\
\bottomrule
\end{tabular}
\caption{Ablation on contrastive augmentation (LLaVA-1.5 backbone). Even without augmentation, \textsc{ReLoop} improves over the base model across all benchmarks; augmentation yields additional gains.}
\label{tab:abl_aug_app}
\end{table}

\section{Case Study}
\label{sec:appendix_case}

We present a qualitative case study to analyze how ReLoop mitigates hallucination across four representative types:

\begin{itemize}[leftmargin=1em]
    \item \textbf{Object Hallucination}: The baseline model incorrectly asserts the presence of a "referee stand" which is not in the image. ReLoop corrects this by recognizing the absence of such an entity.
    \item \textbf{Attribute Hallucination}: An animal is mislabeled as "dog" instead of "chihuahua." ReLoop identifies the finer-grained attribute correctly.
    \item \textbf{Relation Hallucination}: The spatial relationship "on a sofa" is incorrectly predicted; ReLoop grounds the child's location more accurately.
    \item \textbf{Event Hallucination}: The action "not playing" contradicts visual evidence; ReLoop revises the answer to match the depicted motion.
\end{itemize}
As shown in Figure~\ref{fig:case_study}, baseline models such as MiniGPT-4 frequently produce fluent yet inaccurate answers that are not grounded in the image. ReLoop corrects these errors by leveraging consistency feedback to align its answers with both the question intent and visual content. The examples highlight ReLoop's capacity to suppress diverse hallucination patterns and improve factual reliability in open-ended VQA.

\subsection{Handling Nonsensical or Unrelated Initial Answers}

This section complements the above qualitative cases by focusing on an orthogonal failure mode raised by reviewers: how ReLoop handles instances where the initial answer $A$ is nonsensical or unrelated to the question. ReLoop employs a set of structured safeguards: early rejection, $\gamma$-based downweighting via ACW, and entropy-aware masking of attention. The set of structured safeguards can prevent misleading updates when the feedback is deemed unreliable. The representative examples in Table~\ref{tab:case_nonsense} illustrate how these filters operate at the case level before gradients are applied. In addition, the quantitative stress tests in the main text (see Table~\ref{tab:robustness_combined}). demonstrate stable performance under injected noise, corroborating the robustness of these safeguards.

\begin{table*}[t]
\centering
\renewcommand{\arraystretch}{0.95}
\small
\setlength{\tabcolsep}{2pt}

% vertical-centering column types for tight two-column layout
\newcolumntype{L}[1]{>{\raggedright\arraybackslash}m{#1}}
\newcolumntype{C}[1]{>{\centering\arraybackslash}m{#1}}

\begin{tabular}{L{2.8cm} L{3.6cm} L{3.8cm} L{5.0cm}}
\toprule
\textbf{Failure Type} & \textbf{Q} & \textbf{A} & \textbf{Action Taken} \\
\midrule
Empty Output &
``What is on the table?'' &
\emph{(empty)} &
\textit{Early rejection}: sample skipped; no feedback loss computed. \\[2pt]
\addlinespace[2pt]
Overly Generic &
``What sport is being played?'' &
``I'm not sure.'' &
\textit{$\gamma$-downweighting}: low-confidence ACW weight; skipped if below length threshold. \\[2pt]
\addlinespace[2pt]
Nonsensical Repetition &
``What is the man holding?'' &
``Banana banana banana sky help!'' &
\textit{Syntactic abnormality detection}: reject via regex/\#token heuristics; no update. \\[2pt]
\addlinespace[2pt]
Unrelated Semantic &
``What color is the bus?'' &
``Apples grow in the summer.'' &
\textit{Semantic mismatch}: $\gamma\!\approx\!0$ from BERTScore $\Rightarrow$ loss suppressed. \\[2pt]
\addlinespace[2pt]
Hallucinated Words &
``What are the people doing?'' &
``Grockling spinners do fleeb!'' &
\textit{Entropy + token validation}: flat attention masked; spurious tokens filtered before loss. \\
\bottomrule
\end{tabular}
\caption{Case-level handling of nonsensical or unrelated initial answers in ReLoop. Structured safeguards (early rejection, ACW $\gamma$-based downweighting, and entropy-aware masking) prevent misleading gradients from invalid feedback.}
\label{tab:case_nonsense}
\end{table*} 
% \section{Entropy-based Pseudo Ground-Truth Attention $\mathcal{H}_{\text{pseudo}}$}
% \label{appendixC}
% This section includes detailed explanation of the entropy-based mechanism for generating pseudo-ground truth attention maps: Let $A\!\in\!\mathbb{R}^{T\times S}$ denote the raw cross-attention; we row-normalize it to $p_{t,s}=A_{t,s}/\sum_{s'}A_{t,s'}$. For each token $t$, we compute the entropy:
% \begin{equation}
% \mathcal{E}_t \;=\; -\sum_{s=1}^{S} p_{t,s}\log p_{t,s}.
% \end{equation}
% Tokens with high uncertainty (large entropy) are discarded by a threshold $\tau$ (default $\tau{=}2.0$), yielding a confident set $\mathcal{T}_{\text{conf}}=\{\,t\mid \mathcal{E}_t\le\tau\,\}$. The remaining tokens vote for patch importance via attention projection:
% \begin{equation}
% \tilde{h}[s] \;=\; \sum_{t\in\mathcal{T}_{\text{conf}}} p_{t,s}
% \end{equation}
% \begin{equation}
% \mathcal{H}_{\text{pseudo}} \;=\; \mathrm{softmax}\!\big(\tilde{h}/T_a\big)
% \end{equation}
% where $T_a$ is a temperature to avoid over-peaky maps (default $T_a{=}0.7$). We apply light Gaussian smoothing and $\ell_1$ normalization for stability. This mechanism is unsupervised and self-adaptive, based solely on the model’s own attentional confidence, and requires no external saliency detector. 
\section{Entropy-based Pseudo Ground-Truth Attention $\mathcal{H}_{\text{pseudo}}$}
\label{appendixC}

This section details the construction of the entropy-based pseudo ground-truth attention used in Sec.~\ref{sec:4.2.3}. We describe tensor shapes, multi-layer/head aggregation, token filtering, smoothing/normalization, and edge cases to facilitate reproduction.

\paragraph{Notation and Shapes.}
Let the decoder cross-attention at decoding step $t$ be $\{A^{(\ell,h)}_t\in\mathbb{R}^{S}\}_{\ell=1..L,\,h=1..H}$ over $S$ visual patches (keys), for $L$ layers and $H$ heads. We first aggregate heads and layers to obtain a single distribution over patches for token $t$:
\begin{equation}
\bar{a}_t \;=\; \sum_{\ell=1}^L \sum_{h=1}^H w^{(\ell)}\,u^{(h)} \, A^{(\ell,h)}_t
\end{equation}
\begin{equation}
\sum_{\ell} w^{(\ell)}=\sum_h u^{(h)}=1
\label{eq:agg}
\end{equation}
where $w^{(\ell)}$ and $u^{(h)}$ are fixed convex weights. In a default setting, we set uniform across heads ($u^{(h)}{=}1/H$) and a back-loaded layer prior ($w^{(\ell)}\propto \exp(\kappa\,\ell/L)$ with $\kappa{=}1.5$) to emphasize later layers. We row-normalize to obtain a probability over spatial patches:
\begin{equation}
p_{t,s} \;=\; \frac{\bar{a}_t[s]}{\sum_{s'} \bar{a}_t[s']} \;\in\; [0,1],\qquad \sum_s p_{t,s}=1.
\label{eq:norm}
\end{equation}

\paragraph{Per-token Entropy and Confident Set.}
For each generated token $t$, we excludes BOS/EOS/padding/special tokens and computes entropy:
\begin{equation}
\mathcal{E}_t \;=\; -\sum_{s=1}^{S} p_{t,s}\log p_{t,s}\;\;\in\;[0,\log S]
\label{eq:entropy}
\end{equation}
and form a confident token set:
\begin{equation}
\mathcal{T}_{\text{conf}} \;=\; \{\, t \;\mid\; \mathcal{E}_t \le \tau \,\}
\end{equation}
In a default setting, we set $\tau{=}2.0$ (nats). In low-entropy regimes ($\mathcal{E}_t\approx 0$) the attention is highly focused; high entropy indicates diffuse/unstable focus. In an optional variant (ablation only) setting, we reweight votes by $w_t{=}\max\!\big(0,1-\frac{\mathcal{E}_t}{\log S}\big)$; we do not apply this by default to keep the scheme simple and robust.

\paragraph{Voting and Temperature-normalized Map.}
Confident tokens vote for patch importance by summation:
\begin{equation}
\tilde{h}[s] \;=\; \sum_{t \in \mathcal{T}_{\text{conf}}} p_{t,s}
\label{eq:vote}
\end{equation}
We convert $\tilde{h}$ to a soft target by temperature softmax (Default: $T_a{=}0.7$):
\begin{equation}
\mathcal{H}_{\text{pseudo}}[s] \;=\; \frac{\exp\!\big(\tilde{h}[s]/T_a\big)}{\sum_{s'} \exp\!\big(\tilde{h}[s']/T_a\big)}
\label{eq:softmax}
\end{equation}

\paragraph{Spatial Reshaping and Smoothing.}
Let the $S$ patches correspond to a $(H_p \times W_p)$ grid from the vision encoder (e.g., \textit{ViT patch tokens}). We reshape $\mathcal{H}_{\text{pseudo}}\!\in\!\mathbb{R}^{S}$ to $\mathbb{R}^{H_p\times W_p}$, apply light Gaussian smoothing ($3{\times}3$ kernel, $\sigma{=}0.8$; reflect padding), then flatten back to length $S$. A final $\ell_1$ normalization ensures $\sum_s \mathcal{H}_{\text{pseudo}}[s]{=}1$.

\paragraph{Special Tokens, Padding, and Masking.}
We exclude special tokens (BOS/EOS, padding) and punctuation-only tokens from $\mathcal{T}_{\text{conf}}$. For subword tokenization, all subpieces are treated uniformly; no external POS/saliency tools are used to preserve the unsupervised nature.

\paragraph{Empty-set and Degenerate Cases.}
If $\mathcal{T}_{\text{conf}}=\varnothing$ (rare; e.g., \textit{extremely diffuse attention}), we fall back to a min-entropy top-$k$ strategy: pick the $k{=}\max(1,\lceil 0.01T\rceil)$ lowest-entropy tokens to form $\mathcal{T}_{\text{conf}}$ and proceed with Eq.~\eqref{eq:vote}. If $\tilde{h}$ is flat (numerically), we use a near-uniform prior slightly peaked at the global min-entropy token’s argmax.
\begin{algorithm}[t]
{\fontsize{10pt}{12pt}\selectfont
\caption{Constructing Entropy-based Pseudo Attention $\mathcal{H}_{\text{pseudo}}$}
\label{alg:hpseudo}
\begin{algorithmic}[1]
\Require Cross-attn $\{A_{t}^{(\ell,h)}\!\in\!\mathbb{R}^{S}\}$, weights $\{w^{(\ell)}\}$, $\{u^{(h)}\}$, threshold $\tau$, temperature $T_a$
\Ensure $\mathcal{H}_{\text{pseudo}}\!\in\!\mathbb{R}^{S}$ (stop-gradient)
\State $\mathcal{T}_{\mathrm{conf}}\!\gets\!\varnothing$
\For{\textbf{each} token $t$ in generated tokens (exclude specials)}
  \State $\bar{a}\!\gets\!\sum_{\ell=1}^{L}\sum_{h=1}^{H} w^{(\ell)}u^{(h)} A_{t}^{(\ell,h)}$ \Comment{Eq.~\ref{eq:agg}}
  \State $p\!\gets\!\bar{a}/\sum_{s}\bar{a}[s]$ \Comment{Eq.~\ref{eq:norm}}
  \State $\mathcal{E}_{t}\!\gets\!-\sum_{s} p[s]\log p[s]$ \Comment{Eq.~\ref{eq:entropy}}
  \If{$\mathcal{E}_{t}\le\tau$}
    \State $\mathcal{T}_{\mathrm{conf}}\!\gets\!\mathcal{T}_{\mathrm{conf}}\cup\{(t,p)\}$
  \EndIf
\EndFor
\If{$\mathcal{T}_{\mathrm{conf}}=\varnothing$}
  \State $\mathcal{T}_{\mathrm{conf}}\!\gets$ top-$k$ lowest-entropy tokens
\EndIf
\State $\tilde{h}\!\gets\!\sum_{(t,p)\in\mathcal{T}_{\mathrm{conf}}} p$ \Comment{Eq.~\ref{eq:vote}}
\State $\mathcal{H}_{\text{pseudo}}\!\gets\!\mathrm{softmax}(\tilde{h}/T_a)$ \Comment{Eq.~\ref{eq:softmax}}
\State $\mathcal{H}_{\text{pseudo}}\!\gets\!\mathrm{smooth}(\mathrm{reshape\_grid}(\mathcal{H}_{\text{pseudo}}))$
\State $\mathcal{H}_{\text{pseudo}}\!\gets\!\mathcal{H}_{\text{pseudo}}/\|\mathcal{H}_{\text{pseudo}}\|_1$ \Comment{$\ell_1$ norm}
\State \Return $\mathrm{stopgrad}(\mathcal{H}_{\text{pseudo}})$
\end{algorithmic}
}
\end{algorithm}
\paragraph{Objective and Gradient Flow.}
The attention supervision minimizes a KL divergence between the model’s cross-attention map $\mathcal{H}$ (from the current forward pass) and $\mathcal{H}_{\text{pseudo}}$:
\begin{equation}
\begin{aligned}
\mathcal{L}_{\text{attn}}
&= \mathrm{KL}\!\big(\mathcal{H}\,\Vert\,\mathcal{H}_{\text{pseudo}}\big)\\
&= \sum_{s}\mathcal{H}[s]\Big(\log \mathcal{H}[s]-\log \mathcal{H}_{\text{pseudo}}[s]\Big)
\end{aligned}
\end{equation}

\noindent We stop gradients through $\mathcal{H}_{\text{pseudo}}$; only the main model’s attention is updated. The loss is weighted by $\gamma$ in the total objective (Sec.~\ref{sec:acw}).

\paragraph{Pseudocode}
The procedure (Algorithm ~\ref{alg:hpseudo}) is unsupervised and self-adaptive, relying solely on model-internal attentional confidence and requiring no external saliency or human annotation. Its design (entropy gating, temperature control, and mild spatial smoothing) yields stable targets for the KL alignment used in Sec.~\ref{sec:4.2.3}.

\end{document}